\documentclass[sigconf]{acmart}
\AtBeginDocument{%
  }

\usepackage{hyperref}

\usepackage{bm}

\usepackage{subfigure}
\usepackage{soul}
\usepackage{url}

\usepackage{caption}
\usepackage{subcaption}
\usepackage{tablefootnote}
\usepackage{amsmath}   
\usepackage{amsthm}
\usepackage[ruled,linesnumbered]{algorithm2e}

\usepackage{multirow}
\usepackage{diagbox}
\usepackage{float}
\usepackage{array}
\usepackage{flushend}
\usepackage{subcaption}

\usepackage{tablefootnote}
\usepackage{graphicx}
\usepackage{wrapfig}
\usepackage{lipsum} 
\usepackage{colortbl}
\usepackage{balance}
\usepackage{pifont}

\usepackage{diagbox}
\usepackage{multirow}
\usepackage{makecell}
\usepackage{pifont}

\copyrightyear{2026}
\acmYear{2026}
\setcopyright{cc}
\setcctype{by}
\acmConference[WWW '26]{Proceedings of the ACM Web Conference 2026}{April 13--17, 2026}{Dubai, United Arab Emirates}
\acmBooktitle{Proceedings of the ACM Web Conference 2026 (WWW '26), April 13--17, 2026, Dubai, United Arab Emirates}
\acmPrice{}
\acmDOI{10.1145/3774904.3792363}
\acmISBN{979-8-4007-2307-0/2026/04}

\begin{document}

\title{Weighted Graph Clustering via Scale Contraction and Graph Structure Learning}

\author{Haobing Liu}
\affiliation{%
  \institution{Ocean University of China}
  \city{Qingdao}
  \country{China}}
\email{haobingliu@ouc.edu.cn}

\author{Yinuo Zhang}
\affiliation{%
  \institution{Ocean University of China}
  \city{Qingdao}
  \country{China}
}
\email{zyn6261@stu.ouc.edu.cn}

\author{Tingting Wang}
\affiliation{%
 \institution{Ocean University of China}
 \city{Qingdao}
 \country{China}}
\email{wtt2022@stu.ouc.edu.cn}

\author{Ruobing Jiang}
\authornote{Corresponding authors.}
\affiliation{%
  \institution{Ocean University of China}
  \city{Qingdao}
  \country{China}}
\email{jrb@ouc.edu.cn}

\author{Yanwei Yu}
\affiliation{%
 \institution{Ocean University of China}
 \city{Qingdao}
 \country{China}}
\email{yuyanwei@ouc.edu.cn}
\authornotemark[1]

\renewcommand{\shortauthors}{Haobing Liu, Yinuo Zhang, Tingting Wang, Ruobing Jiang \& Yanwei Yu}

\begin{abstract}
Graph clustering aims to partition nodes into distinct clusters based on their similarity, thereby revealing relationships among nodes. Nevertheless, most existing methods do not fully utilize these edge weights. Leveraging edge weights in graph clustering tasks faces two critical challenges. (1) The introduction of edge weights may significantly increase storage space and training time, making it essential to reduce the graph scale while preserving nodes that are beneficial for the clustering task. (2) Edge weight information may inherently contain noise that negatively impacts clustering results. However, few studies can jointly optimize clustering and edge weights, which is crucial for mitigating the negative impact of noisy edges on clustering task. To address these challenges, we propose a contractile edge-weight-aware graph clustering network. Specifically, a cluster-oriented graph contraction module is designed to reduce the graph scale while preserving important nodes. An edge-weight-aware attention network is designed to identify and weaken noisy connections. In this way, we can more easily identify and mitigate the impact of noisy edges during the clustering process, thus enhancing clustering effectiveness. We conducted extensive experiments on three real-world weighted graph datasets. In particular, our model outperforms the best baseline, demonstrating its superior performance. Furthermore, experiments also show that the proposed graph contraction module can significantly reduce training time and storage space.
\end{abstract}

\ccsdesc[500]{Computing methodologies~Unsupervised learning}
\ccsdesc[500]{Computing methodologies~Neural networks}

\keywords{Graph Clustering, Graph Attention Network, Graph Structure Learning, Graph Contraction}

\maketitle

\section{Introduction}
Graph clustering~\cite{liu2023hard,wen2023efficient}, a specialized form of clustering, aims to partition nodes within a graph into distinct clusters based on their connection patterns. Its primary objective is to ensure high similarity among nodes within the same cluster while minimizing similarity between nodes in different clusters. Graph clustering plays a crucial role in uncovering relationships among nodes in a graph. It has numerous applications in both academia and industry, including community detection~\cite{srichandra2024community}, image recognition~\cite{zhong2021graph}, and recommendation systems~\cite{chunjing2025learning}.

Existing deep graph clustering methods can be categorized into two classes based on their network architecture: graph neural network (GNN)-based and multi-layer perceptron (MLP)-based methods. MLP-based methods~\cite{yang2022contrastive,liu2023deep} utilize MLPs to extract informative features from graph data. But they struggle to effectively capture non-euclidean structural information. Recent progress in the Web domain has further validated the role of structural integration in deep clustering~\cite{DBLP:conf/www/Bo0SZL020} and the use of advanced graph representations~\cite{DBLP:conf/www/BoutalbiBVS24} in facilitating the capture of intricate user patterns.
In recent years, non-Euclidean graph data has become increasingly pervasive, as it encapsulates not only node attributes but also topological structure information, which characterizes the relationships among data points~\cite{DBLP:conf/icml/Pan023}. To fully leverage this information, GNN-based methods~\cite{jin2019graph,bhowmick2024dgcluster} employ GNN encoders (e.g., Graph Convolutional Network (GCN)~\cite{DBLP:conf/iclr/KipfW17}, Graph Attention Network (GAT)~\cite{velickovic2017graph}) to model graph data. These methods propagate information by aggregating features from neighboring nodes and learning low-dimensional representations, which brings nodes within the same cluster closer in the embedding space. Although some methods can handle weighted graphs (e.g., by introducing adjacency matrices with edge weights), almost all of them neglect that edge weight information often contains noise. 

\begin{figure}[t]
\centering 
\includegraphics[width=4.5cm]{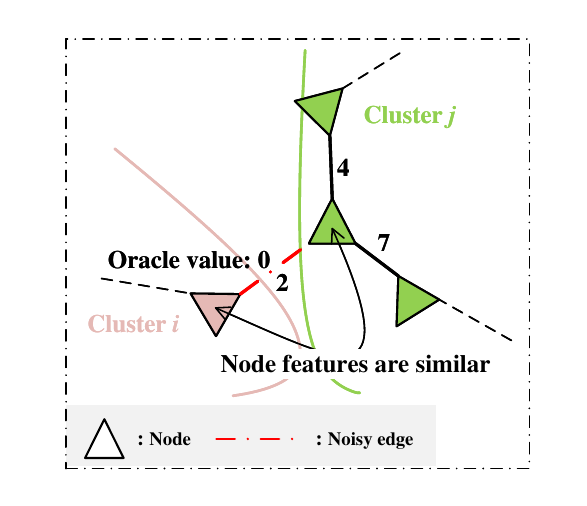}  
\vspace{-2mm}
\caption{An illustrative example of graph clustering on a noisy weighted graph.}
\label{intro}
\vspace{-2mm}
\end{figure}

Using edge weights effectively in graph clustering faces two critical challenges. \textit{Firstly,} the introduction of edge weights significantly increases resource cost. Tuples are often chosen to store the graph structure due to the requirement of scalability. If edge weights are not considered, 2-tuples (i.e., (node, node)) can be leveraged. While edge weights are taken into account, 3-tuples (i.e., (node, node, weight)) should be utilized. This leads to a 50\% increase in GPU memory usage during the model training process. Furthermore, this may also increase the time cost. To mitigate this issue, subgraph sampling methods~\cite{zou2019layer,zhang2023subgraph} are adopted to extract a subgraph from the original graph, thereby reducing costs. However, existing sampling strategies often randomly select nodes or edges, which can lead to subgraphs that are not sufficiently representative of the original graph's clustering structure. Random sampling may exclude nodes that are densely connected within a cluster (they are pivotal for defining internal cohesion). If density is considered during sampling, nodes in low-density regions may be overlooked, causing the number of clusters in the subgraph to be lower than the true value. 
Therefore, a critical challenge lies in \textit{how to reduce the scale of weighted graphs while preserving nodes that contribute significantly to the clustering structure.}

\textit{Secondly,} graph structures often contain noisy edges~\cite{you2020handling,wang2021graph}, indicating a discrepancy between the edge weights and their corresponding oracle values. Such noisy edge weights may lead target nodes to aggregate more features from unrelated neighbors, thereby leading to incorrect clustering results. As shown in Figure~\ref{intro}, node features of the green and red nodes are similar, and they are connected by an edge with a weight of 2. Existing methods might cluster these two nodes into the same group. In reality, these two nodes belong to distinct clusters. Some methods~\cite{sun2022graph,he2023graph} attempt to mitigate the impact of noisy edges by optimizing adjacency matrices. However, these methods rarely optimize graph structure with clustering together.
Consequently, a critical challenge is \textit{how to jointly optimize clustering and edge weights, thereby removing noisy edges that harm clustering task.}

To address these challenges, we propose a \underline{c}ontractile \underline{e}dge-w\underline{e}ight-aware \underline{g}raph \underline{c}lustering \underline{n}etwork (CeeGCN in short). To solve the \textit{first challenge}, we design a cluster-oriented graph contraction module. This module extracts a subgraph from the original graph by jointly considering node density, distance, and similarity. The extraction process preserves the intrinsic clustering structure of the original graph. To tackle the \textit{second challenge}, we design an edge-weight-aware sparse graph attention network (EWSGAT in short). Compared to GAT, EWSGAT explicitly integrates both node features and edge weights when computing attention coefficients. It then utilizes an $\alpha$-entmax function~\cite{peters2019sparse} for normalization, which automatically sparsifies attention coefficients by effectively setting the weights of noisy edges to zero, thereby suppressing the influence of noisy edges. Based on sparse attention coefficients, we can optimize the graph structure, thereby improving clustering performance.

For clarity, the main contributions of this paper are summarized as follows:
\begin{itemize}
\item We design a cluster-oriented graph contraction module that significantly reduces resource consumption during training while ensuring clustering performance. This lays the foundation for future solutions to large-scale weighted graph clustering problems.
\item We develop an edge-weight-aware sparse graph attention network that reduces the impact of noisy edges and can optimize the graph structure. This ensures our model with robustness and scalability.
\item We construct three weighted graph datasets in the real world. We conduct extensive experiments on these datasets. The experimental results demonstrate that the proposed model (i.e., CeeGCN) outperforms competitive methods. In particular, CeeGCN with an average Micro-F1 improvement of 31.5\% compared with the best baseline.
\end{itemize}
\vspace{-6mm}

\section{Related Work}
\subsection{GNN-based Graph Clustering} 
In recent years, graph clustering techniques have demonstrated promising performance~\cite{kang2024cdc}. In particular, with the help of GNNs, graph clustering methods can leverage both node features and graph structures. GNN-based methods utilize GNNs to obtain node representations by aggregating features from neighbors, subsequently clustering these representations. DGCluster~\cite{bhowmick2024dgcluster} proposes a trainable modular method that models the graph structure using GNN (e.g., GCN, GAT) to learn node representations, thereby accomplishing cluster number estimation. Some methods introduce GCNs for learning node representations. For example, SSGCN~\cite{he2023graph} employs a graph structure learning layer and an adaptive graph convolutional layer to jointly optimize both graph structure and node representations. DyFSS~\cite{zhu2024every} assigns each self-supervised learning (SSL) task to a task-specific GCN layer to extract features using the corresponding SSL loss.

Some methods introduce GATs, empowering models to dynamically learn the importance of neighbors. NCAGC~\cite{wang2022ncagc} introduces a neighborhood contrastive mechanism to directly construct positive and negative sample pairs within the node representation space generated by a GAT encoder. DNENC~\cite{wang2022deep} combines a dual-branch autoencoder to simultaneously reconstruct both structural and attribute information. EFR-DGC~\cite{hao2023deep} utilizes a GAT to construct a graph autoencoder to extract the topological relationships of nodes. This information is then combined with hierarchical attribute information to generate enhanced node representations that are beneficial for clustering.

Additionally, some methods~\cite{huo2023caegcn} integrate GCNs with autoencoders to fully exploit node features. SCGC~\cite{kulatilleke2025scgc}, based on graph autoencoders, aims to improve the efficiency and scalability of existing deep graph clustering methods by decoupling transformation and propagation operations.

\subsection{Graph Sampling and Graph Structure Learning}
Existing graph sampling methods fall into three classes based on their sampling granularity: node sampling, layer-wise sampling, and subgraph sampling~\cite{zou2019layer,zhang2023subgraph}. Node sampling methods~\cite{,zhang2022hierarchical,shin2023efficient} randomly select a subset of nodes from a target node's neighborhood for message passing. Layer-wise sampling methods~\cite{zou2019layer,balin2023layer} divide the graph into multiple layers, then perform global sampling at each message passing layer. However, both classes of methods require resampling before each forward pass. Subgraph sampling methods~\cite{zeng2019graphsaint,zhang2023subgraph} sample relatively independent connected subgraphs from the original graph, training each subgraph independently as a batch. Yet current random selection strategies often compromises global structure and fails to preserve critical edges.

\begin{figure*}[t]
    \centering
    \includegraphics[scale=0.37]{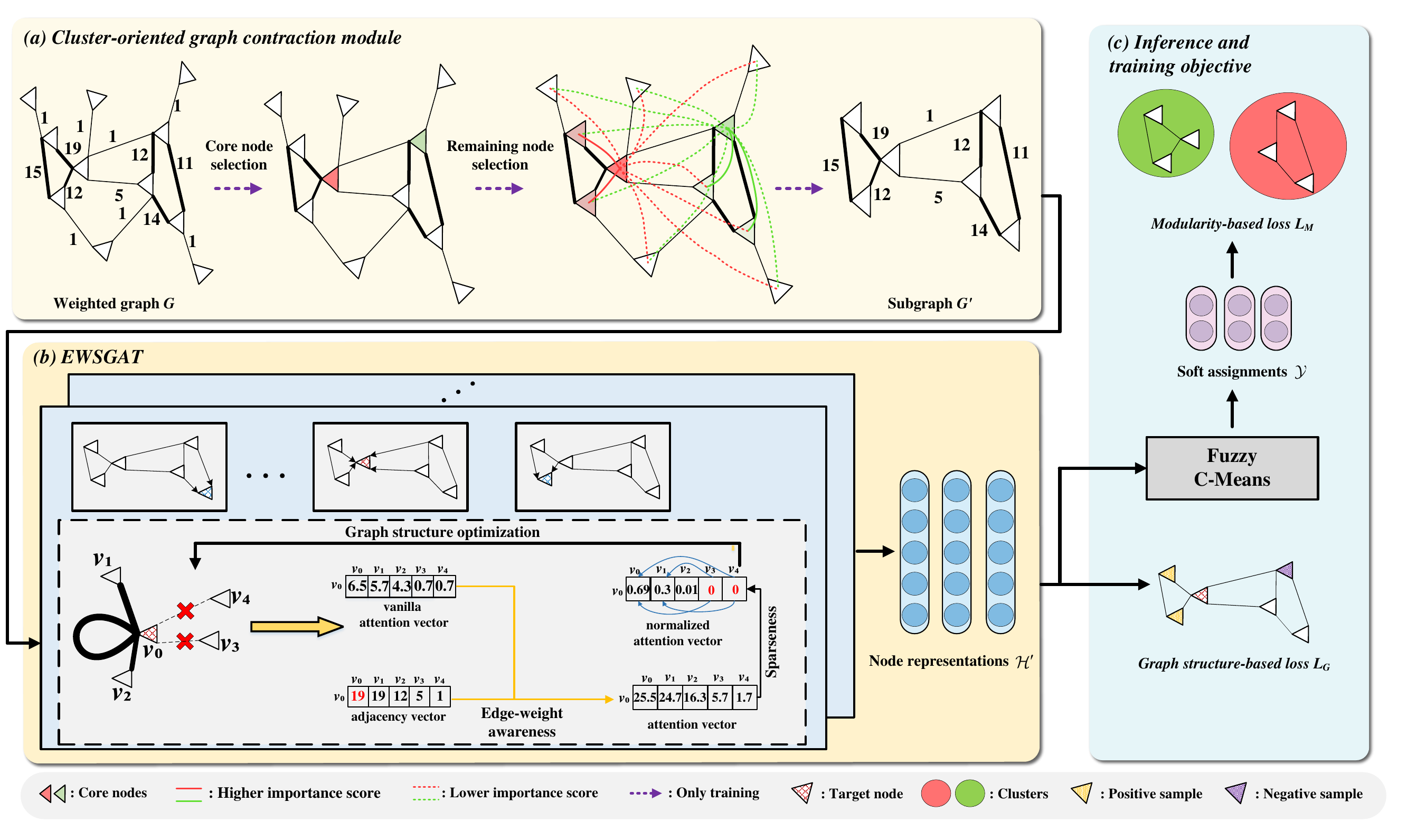} 
    \vspace{-3mm}
    \caption{The architecture of CeeGCN, including (a) Cluster-oriented graph contraction module, (b) EWSGAT, and (c) Inference and training objective.} 
    \vspace{-4mm}
    \label{model}
\end{figure*}

The core idea of graph structure learning is to jointly optimize the graph structure itself along with node representations. VIB-GSL~\cite{sun2022graph} generates an information bottleneck graph through feature masking and graph structure learning and then use GNNs to encode this graph to obtain node representation. SSGCN~\cite{he2023graph} introduces a graph structure learning layer before each graph convolution layer, exploring end-to-end joint graph structure and representation learning. However, these methods overlook edge weights and can not jointly optimize graph structure and clustering. Specifically, since nodes connect to both homophilous (class-consistent) and heterophilous (class-divergent) neighbors~\cite{shen2025heterophily}, the lack of refined modeling of edge weights hinders effectively suppress noise introduced by heterophilous neighbors during aggregation process.

\section{Preliminaries}
\textbf{Weighted Graph:} A weighted graph is defined as $G=(V,E,\mathcal{\bm{W}})$, $V=\{v_1,v_2,\cdots,v_n\}$ is the set of node, where $n$ is the number of nodes, and $E$ is the set of edges. $\mathcal{\bm{W}}$ is the weight matrix. In an undirected weighted graph, if there is an edge between $v_i$ and $v_j$, $w_{ij} \ne 0$; otherwise, $w_{ij} = w_{ji}=0$. \\
\textbf{Problem Formulation:} Given a weighted graph $G=(V,E,\mathcal{\bm{W}})$, the goal of graph clustering is to partition the $n$ nodes in the graph into $K$ disjoint clusters $C=\{C_1,C_2,\cdots,C_K\}$ so that each node $v_i\in V$ is assigned a cluster label $c_i \in              \{1,2,\cdots,K\}$ indicating its membership. $\forall k, k^{\prime}$, $C_k \cap C_{k^{\prime}}=\emptyset$. \\
\textbf{Core nodes:} Core nodes are a small set of representative nodes jointly selected based on two criteria: local density and distance between nodes. These nodes have high local density and cover a wide range of the original graph. We will introduce the details of the selection in Section~\ref{cgc}.\\
\textbf{Subgraph:} The subgraph $G^{\prime}$ is a subset of nodes and edges selected from the original graph. In this paper, a cluster-oriented graph contraction module is designed to get $G^{\prime}$. Specifically, core nodes are first selected; then the remaining nodes are selected based on their importance relative to the core nodes. These nodes and the edges among them form $G^{\prime}$.\\
\vspace{-2em}

\section{The Proposed Method}
The overall architecture of our proposed model (i.e., CeeGCN) is shown in Figure~\ref{model}. The model consists of three modules: (a) Cluster-oriented graph contraction module. This module first selects core nodes and important nodes around the core nodes, and constructs a subgraph. (b) EWSGAT. This module calculates attention coefficients by considering both node features and edge weights. Noise edges are suppressed via the $\alpha$-entmax function. (c) Inference and training objective. The training process is jointly guided by a graph structure-based loss and a modularity-based loss.

\subsection{Cluster-oriented Graph Contracting}{\label{cgc}}
As shown in Figure~\ref{model}(a), we propose a cluster-oriented graph contraction module that selects core nodes by combining node density and distance, and the remaining nodes are selected based on their importance relative to the core nodes.

Firstly, we introduce the selection of core nodes. High-density nodes typically reside at the core of clusters. Meanwhile, core nodes should maintain appropriate spacing to maximize coverage of the entire graph. 

We estimate the density $\rho_i$ of node $i$: $\rho_i = \sum_{j \in \mathcal{N}_i} w_{ij}$, where $w_{ij}$ denotes the weight between node $i$ and node $j$, and $\mathcal{N}_i$ denotes the set of neighbors of node $i$. The node with the highest density is selected as the initial core node.

Assume that we should select $O$ core nodes. To determine the remaining $O-1$ core nodes, we estimate the distance between node $i$ and selected core nodes: $\theta_i = \sum_j \mathrm{dist}(v_i,\ v_j)$, where $v_j$ represents a selected core node; $\mathrm{dist}(v_i, v_j)$ denotes the shortest path length from node $i$ to node $j$, calculated using the Dijkstra algorithm. Let $\Re _{\rho _{i}} = \mathrm{rank}\left ( \rho _{i}  \right ) $ and $\Re _{\theta  _{i} } =\mathrm{rank}\left ( \theta  _{i}  \right ) $ denote the scores obtained by sorting the respective values in ascending order, where nodes with earlier-ranked positions are assigned higher scores. Then comprehensive scores are obtained through the weighted integration of $\Re _{\rho _{i}}$ and $\Re _{\theta  _{i} }$:
\begin{equation}
\begin{aligned}
\label{equ10}
\Re _{i} =\varepsilon \Re _{\rho _{i} } +\left ( 1-\varepsilon  \right ) \Re _{\theta  _{i} },
\end{aligned}
\end{equation}
where $\varepsilon$ is a hyperparameter that balances the influence of density and distance. The node with the highest comprehensive score is selected as a core node. This selection process is iterated a total of $O-1$ times.

Secondly, we introduce the selection of remaining nodes. We calculate the importance scores from each core node to other nodes via the personalized PageRank algorithm~\cite{gupta2008fast}:
\begin{equation}
{\bm{S} = \varphi \cdot ( \mathcal{\bm{I}} - (1 - \varphi) \cdot \mathcal{\bm{W}} \mathcal{\bm{D}}^{-1} )},
\end{equation}
where $\mathcal{\bm{I}}$ is the identity matrix, $\mathcal{\bm{D}}$ is the degree matrix and $\varphi$ is a tunable hyperparameter. 
Based on $S_i$ (where $i$ denotes the core node $i$), nodes with important scores greater than a threshold are selected.  

All selected nodes and the edges among them form a subgraph $G^{\prime}$, which captures the original graph's clustering structure.

\subsection{Edge-weight-aware Sparse Graph Structure Learning}
Vanilla GAT computes attention coefficients solely based on node features, neglecting valuable edge weight information. In this paper, we propose an edge-weight-aware sparse graph attention network (i.e., EWSGAT), as shown in Figure~\ref{model}(b), which explicitly incorporates edge weights when calculating attention coefficients. The attention coefficient is redefined as follows:
\begin{equation}
\label{zong}
\begin{aligned}
e_{iz}^\prime={f_{iz}}+{e_{iz}},\ z\in{\mathcal{N}_i}\cup v_i
\end{aligned}
\end{equation}
where $e_{iz}$ denotes the vanilla attention coefficients; $f_{iz}$ is the impact factor that is derived from edge weights; ${\mathcal{N}_i}$ denotes the set of neighbors for node $i$. $e_{iz}$ is calculated as follows:

\begin{equation}
e_{iz}=({\bm{W_1}{\bm{{h}_i}}})^\top\cdot(\bm{W_1}{\bm{{h}_z}}),
\end{equation}
where $\bm{W_1}$ denotes a learnable weight matrix. $\bm{{h}_i}$ and $\bm{{h}_z}$ are node representations of node $i$ and node $z$, respectively, which serve as the input to the EWSGAT layer. $f_{iz}$ is calculated as follows:

\begin{equation}
\label{quan}
\begin{aligned}
f_{iz}=\frac{w_{iz}}{\sum\limits_{j\in {\mathcal{N}_i}}w_{ij}+w_{ii}},
\end{aligned}
\end{equation}
where $w_{ii}$ represents the self-loop weight of node $i$. $w_{ii}=\mathrm{max}\ {\{w_{ij}\}}$, $j\in{\mathcal{N}_i}$. 

Compared with the vanilla calculation method, higher edge weights lead to larger attention coefficients. After normalization, these larger attention coefficients will be further significantly increased.

To normalize the attention coefficients, we further introduce the $\alpha$-entmax function to replace the softmax function used in vanilla GAT. By incorporating the hyperparameter $\alpha$, when $1<\alpha<2$, any attention score below a certain threshold is set to 0, focusing the attention on the most related neighbors and reducing the impact of noisy connections.

Formally, we collect the redefined attention coefficients for all neighbors of node $i$, denoted as $\bm{e_{i}^{\prime}} = [e_{i1}^{\prime}, e_{i2}^{\prime}, \dots, e_{ij}^{\prime}, \dots e_{i({\lvert\mathcal{N}_i\rvert+1)}}^{\prime}]$, where $\mathcal{N}_i$ denotes the set of neighbors of node $i$; $|\bm{e_{i}^{\prime}}|=|\mathcal{N}_i \cup v_i|=\lvert\mathcal{N}_i\rvert+1$. The normalized redefined attention coefficients between node $i$ and its neighbors are then obtained by:
\begin{equation}
\bm{{a}_i^{\prime}} = \mathrm{{\alpha}\text{-}entmax}(\bm{e_{i}}^\prime)={[(\alpha-1)\bm{e_{i}}^\prime-\tau \mathbf{\textbf{1}}]_+^\frac{1}{\alpha - 1}},
\end{equation}
where $\mathbf{\textbf{1}}$ represents a vector of all ones, $[x]_+=\mathrm{max}\{0,\; x\}$, and by $[\bm{x}]_+$ its elementwise application to vectors. So $[(\alpha-1){e_{iz}}^\prime-\tau]_+=\mathrm{max}\{0,\; [(\alpha-1){e_{iz}}^\prime-\tau]\}$. $\tau$ denotes the threshold for $\alpha$-entmax. When the attention coefficient satisfies ${e_{iz}^{\prime}} \leq \frac{\tau}{\alpha - 1}$, the normalized probability becomes 0.

To ensure that all probabilities sum to 1, $\tau$ must satisfy:
\begin{equation}
\begin{aligned}
{\sum\limits_{j}[(\alpha-1){e_{iz}^{\prime}}-\tau({e_{iz}^{\prime}})]_+^{\frac{1}{\alpha - 1}}=1},
\end{aligned}
\end{equation}
where $\tau({e_{iz}^{\prime}})$ denotes the dynamically computed the value of $\tau$ based on ${e_{iz}^{\prime}}$. 
Considering that each node may have a large number of neighbors, we adopt the binary search method~\cite{ref18} to calculate $\tau$. The detailed calculation process is provided in the Appendix.

After obtaining the normalized attention coefficients, the node representation can be updated using representations of its neighbors as follows:
\begin{equation}
\begin{aligned}
{\bm{{h}_i}}^{\prime}= \sigma(\sum_{z \in {\mathcal{N}_i \cup v_i}} {a}_{iz}^{\prime}\bm{W_2}\bm{{h}_z}),
\label{equ14}
\end{aligned}
\end{equation}
where $\sigma(\cdot)$ is the ELU~\cite{clevert2015fast} function; ${a}_{iz}^{\prime}$ is the normalized attention coefficient between node $i$ and node $z$; $\bm{W_2}$ is a learnable weight matrix; $\bm{{h}_i}^{\prime}$ is the node representation of node $i$, which serve as the output of the EWSGAT layer.

Vanilla multi-head attention mechanism may neglect critical information from certain attention heads, so we introduce a learnable parameter $\gamma_{t}$ to adaptively assign importance to head $t$. Assume that there are $T$ heads. We fuse these heads to obtain the final node representation of node $i$ as follows:
\begin{equation}
\begin{aligned}
{\bm{{h}_i}^{\prime}} = \sum_t{\gamma_{t} \bm{{h}_{it}}'},\ t = 1, 2,\cdots, T,
\label{finaleq}
\end{aligned}
\end{equation}
where $\gamma_{t}$ is a learnable parameter and $\bm{{h}_{it}'}$ can be calculated with Eq.~\eqref{equ14}. 
Finally, all node representations output by EWSGAT can be concatenated to form a node representation matrix $\mathcal{\bm{H}'}$.

\begin{algorithm}[t]
	\caption{Training and inferencing of CeeGCN.}
        \label{algorithm1}
        
            \KwIn{the original graph $G$, the number of clusters $K$, and hyperparameters}
            \KwOut{$K$ disjoint clusters $C$}
            \BlankLine
            Select the first core node via node density: $\rho_i = \sum_{j \in \mathcal{N}_i} w_{ij}$;
            \BlankLine

            \Repeat{selecting $O-1$ additional core nodes}{
            Compute distance $\theta_i = \sum_j \mathrm{dist}(v_i, v_j)$;\\
            Select core node with the highest comprehensive score $\Re_i$ based on density and distance;
            }
            Calculate importance scores: $S = \varphi \cdot ( \mathcal{I} - (1 - \varphi) \cdot \mathcal{W} \mathcal{D}^{-1} )$ and select the nodes with scores greater than a threshold as the remaining nodes; 
             
            Construct subgraph $G^{\prime}$ based on selected nodes;     
    	
             Initialize the node representations matrix $\mathcal{H}$;	
    	
            \Repeat{convergence criterion is satisfied}
            {
                $\mathcal{\bm{H}}'=\mathrm{EWSGAT}(\mathcal{H})$;

                $\mathcal{Y}={\mathrm{FCM}}(\mathcal{\bm{H}}')$;
                
                Assign clusters: $c_i={\mathrm{{\underset{{\textit{k}}\in\{1,2,\cdots,K\}}{argmax}}}y_{ik}}$;  
                
                Updated edge weights: $w_{ij}\leftarrow\frac{{a}_{ij}^{\prime}+{a}_{ji}^{\prime}}{2}\cdot w_{ij}$;
                
                Compute the total loss $L$ via $L = L_G+\eta{L_M}$;
                
                Backpropagate and update model parameters using $L$;
    	}	
            
      Input the original graph $G$ into the trained model to obtain the prediction result $C$.
\end{algorithm}
\subsection{Inference and Training Objective}

To obtain soft clustering assignments, we apply the Fuzzy C-Means (FCM) clustering algorithm~\cite{bezdek1981objective} to $\mathcal{\bm{H}'}$, as shown in Figure~\ref{model}(c). Unlike the standard euclidean distance, our approach directly utilizes the inner product to characterize the relationship between nodes and cluster centers. Specifically, the optimization process is performed iteratively. In each iteration, we first update the membership probability $\mathcal{Y}_{ik}$ of node $i$ belonging to cluster $k$ via:
\begin{equation} \mathcal{Y}_{ik} = \left[ \sum_{j=1}^K \left( \frac{\bm{{h}_i}^{\prime \top} \bm{\mu}_k}{\bm{{h}_i}^{\prime \top} \bm{\mu}_j} \right) \right]^{-1}, \end{equation}
where $\bm{\mu}_k$ denotes the vector of the $k$-th cluster center. Each center $\bm{\mu}_k$ is computed as a weighted average of the node representations using its corresponding membership probability $\mathcal{Y}_{ik}$ as weights. The final cluster labels $c_i$ are determined by the index of the maximum membership probability.

In weighted graphs, modularity calculation depends not only on node degrees but also requires a comprehensive consideration of the edge weights to reflect the strength of the connections between nodes. To adapt to the dynamically learned graph structure, we replace the original edge weights with a combination of the original edge weights and the learned attention coefficients. This allows for a more accurate modeling of the structural relationships within the current graph. The graph structure optimization between node $i$ and node $j$ is performed as follows:
\begin{equation}
\begin{aligned}
w_{ij}\leftarrow\frac{{a}_{ij}^{\prime}+{a}_{ji}^{\prime}}{2}\cdot w_{ij}.
\end{aligned}
\end{equation}

The modularity-based loss function is expressed as follows:
\begin{equation}
\begin{aligned}
L_M = -\frac{1}{2m}{\sum_{i,j}}(w_{ij}-{\frac{k_ik_j}{2m}})\delta(c_i,c_j),
\end{aligned}
\end{equation}
where $m=\frac{1}{2}{\sum_{ij}}w_{ij}$ denotes the sum of the weights of all edges, $k_i={\sum_{j}}w_{ij}$ denotes the sum of the weights of all edges connected to the node $i$, and $c_i$ denotes the $i$-th cluster, $\delta(c_i, c_j)=1$ when $c_i=c_j$, otherwise 0.

To enhance the discriminability of node representations, we introduce a loss based on contrastive learning. Specifically, for each target node $i$, we select its neighbors as positive samples to reconstruct the graph structure, while selecting unrelated nodes as negative samples. The graph structure-based loss function of node $i$ can be expressed as:
\begin{equation}
\begin{aligned}
L_i=\mathrm{log}(\sigma(\bm{{h}_i}^{\prime}{}^\top\bm{{h}_u}^{\prime}))-Q\times\mathbb{E}_{v\sim{P_n(u)}}\mathrm{log}(\sigma({-\bm{{h}_i}}^{\prime}{}^\top{\bm{{h}_{v}}}^{\prime})),
\end{aligned}
\end{equation}
where $u$ is the positive sample, $P_n$ is the probability distribution of negative samples, $Q$ is the number of negative samples, and $\sigma(\cdot)$ denotes the sigmoid function. 

The overall graph structure-based loss is defined as the average of individual node-level losses across the entire graph, formulated as $L_G = \frac{1}{n}{\sum_{i=1}^n}{L_i}$.

The final optimization objective function combines the above two sub-loss functions in a weighted manner:
\begin{equation}
\begin{aligned}
L = L_G+\eta{L_M}.
\label{equ21}
\end{aligned}
\end{equation}
where $\eta$ is a hyperparameter. 

All parameters are learned based on the subgraph. After training is complete, the original graph is sent to the trained model (without passing through cluster-oriented graph contraction module) to obtain clustering results. The entire procedure is shown in Algorithm~\ref{algorithm1}.

\begin{table}[b]
\setcounter{table}{0}
\begin{center}
\small

{\caption{Dataset statistics.}\label{table1}}

\begin{tabular}{l|r|r|c|c}
\toprule
\textbf{Dataset} & \multicolumn{1}{c|}{\textbf{Nodes}} & \multicolumn{1}{c|}{\textbf{Edges}} & \textbf{Clusters} & \textbf{Density} \\ 
\midrule
Vessel01        & 818                                 & 144,898                            & 4                    & 0.434           \\
Vessel10        & 763                               & 146,981                             & 4                    & 0.505            \\
ML100k      & 1,612                               & 58,424                              & 9                    & 0.045            \\ 

\bottomrule
\end{tabular}
\end{center}
\end{table}

\begin{table*}[]
\setcounter{table}{1}
\setlength{\tabcolsep}{3.2pt} 
\centering 
\small
\caption{Clustering performance on five datasets. The bold and underlined values indicate the best and the runner-up results.}\label{baseline}
\setlength{\tabcolsep}{10pt}
\begin{tabular}{l|ccl|cc|cc|cc|cc}
\toprule
\multirow{2}{*}{\textbf{Model}} & \multicolumn{3}{|c|}{\textbf{Vessel01}}  & \multicolumn{2}{|c|}{\textbf{Vessel10}} & \multicolumn{2}{|c|}{\textbf{ML100K$_{N=0\%}$}} & \multicolumn{2}{|c|}{\textbf{ML100K$_{N=10\%}$}} & \multicolumn{2}{|c}{\textbf{ML100K$_{N=20\%}$}} \\
\cmidrule(lr){2-12} 
                                & ACC              & \multicolumn{2}{c|}{F1}    & ACC           & F1             & ACC                & F1                 &       ACC     &   F1         &     ACC       &   F1         \\ \midrule
SSGCN                           & 43.15            & \multicolumn{2}{c|}{29.49} &     44.88          & 26.25          & 37.66              & 16.57              &   36.48         &     15.15       &  36.41          &    14.43        \\
DyFSS                           & 53.06            & \multicolumn{2}{c|}{30.50} &       52.93        & 30.44          & 24.38              & 13.54              &     23.88       &     13.07       &     23.70       &    12.87        \\
DGCLUSTER                       & 58.67            & \multicolumn{2}{c|}{57.58} &       51.11        & 56.60          & 36.84              & 32.85              &    36.35        &     37.14       &    35.85        &   35.28         \\
SCGC                            & 56.50            & \multicolumn{2}{c|}{35.43} &     48.62          & 26.90          & 44.57              & 17.38              &     44.04       &     14.00       &          \underline{43.73 } &     14.06       \\ 
DAGC                            & 50.93           & \multicolumn{2}{c|}{32.02} &     59.13          & 23.26          & 33.91              & 10.96              &      21.99      &     12.32      & 21.54         &   11.13        \\
GC-SEE                           & 38.84            & \multicolumn{2}{c|}{23.91} &     32.06          & 21.17          & 36.43              & 31.36              &   33.96         &     25.39       &  33.17         &   22.14        \\
VGAER                           & 32.32            & \multicolumn{2}{c|}{34.43} &     33.95          & 36.46          & 43.44              & \underline{48.48}              &   36.85        &     \underline{42.32}       &  37.21          &    32.58        \\
SUBLIME                           & 31.73  & \multicolumn{2}{c|}{20.49} &       32.01        & 21.17          & 20.83              & 12.39              &     22.05       &     11.69       &     22.02       &    12.64        \\
NMFGAAE                       & 36.35            & \multicolumn{2}{c|}{28.93} &       30.52        & 24.38          & 35.64              & 40.19              &   31.71        &     35.38       &    34.06        &   38.73         \\
CaEGCN                            & 58.28           & \multicolumn{2}{c|}{36.38} &     53.43          & 35.70          &44.21             & 45.43              &     44.19       &     \textbf{44.71}       &    43.15 &     \underline{44.52}      \\ 
GMIM                            & \underline{78.61 }           & \multicolumn{2}{c|}{\underline{69.51}} &     \underline{74.84}          & \underline{65.75}          & \underline{44.78}              & 31.17              &      \underline{44.75}      &      28.17      &  \textbf{43.79 }         &    29.81        \\
\midrule
CeeGCN                          & \textbf{78.97}            & \multicolumn{2}{c|}{\textbf{88.25}} &   \textbf{ 75.36}           & \textbf{85.94}          & \textbf{46.36}              & \textbf{63.35}              &    \textbf{44.97}        &     41.03       &  43.64         &  \textbf{60.76}
         \\ 
         \bottomrule
\end{tabular}
\end{table*}

\section{Experiments}
\noindent\textbf{Datasets.} 
The proposed model in this paper is evaluated on three datasets, and the statistical information for each dataset is presented in Table~\ref{table1}. Node features only include node ID. If other features were available, they could also be included.

Vessel01 and Vessel10 are constructed from the vessel trajectory data from four cities, including timestamps, longitude and latitude positions, and heading direction.

\begin{itemize}
\item When the spherical distance between two vessels is less than 1.11 kilometers, they are considered to meet, creating an edge with a weight increment of one.
\item As vessels belonging to the same city frequently encounter each other, we use the city to which a vessel belongs as the ground label.
\item Each dataset represents the encounters of valid vessel trajectories for a specified month.
\end{itemize}

ML100K is constructed from the dataset MovieLens 100K. We will include this content in the Supplementary Material. The dataset includes multiple files, with our main focus being on u.data and u.item. (We will release this dataset.)

\begin{itemize}
\item Constructing a graph from rating records of users in u.data. First, filter the rating records by each user and sort them by timestamp. Then, compute the edge weights based on consecutive movie ratings. If two movies are rated consecutively by the same user, the edge weight between these two movies is increased by 1.
\item Assign labels to movies based on genres in u.item. Since movies may belong to multiple genres, we calculate the frequency of each genre across the dataset and assign the label based on the genre with the highest frequency for each movie.
\end{itemize}

\noindent\textbf{Methods for Comparison.} 
We compare our proposed method with eleven competitive methods to evaluate the effectiveness of our method, including \textbf{SSGCN}~\cite{he2023graph}, \textbf{DyFSS}~\cite{zhu2024every}, \textbf{DGClUSTER}~\cite{bhowmick2024dgcluster}, \textbf{SCGC}~\cite{kulatilleke2025scgc}, \textbf{DAGC}~\cite{ref29}, \textbf{GC-SEE}~\cite{ref28}, \textbf{VGAER}~\cite{ref30}, \textbf{SUBLIME}~\cite{liu2022towards}, \textbf{NMFGAAE}~\cite{ref31}, \textbf{CaEGCN}~\cite{ref33}, and \textbf{GMIM}~\cite{ahmadi2024deep}.

\noindent\textbf{Evaluation Metrics and Experimental Setting.} 
We select two widely used metrics to evaluate the clustering results: Accuracy (ACC) and Micro F1-score (F1). We multiply the metrics by 100 for better readability. Our method is implemented with the PyTorch 2.3.0\footnote[1]{\url{https://github.com/HaobingLiu/CeeGCN}}. For baselines, we download the source codes and report the corresponding clustering results using the original paper's settings. For our proposed method, the number of heads is set as 8, and the number of EWSGAT layers is set to 3. We tune ``$\varepsilon$'' in the calculation of comprehensive scores, ``$\alpha$'' in the $\alpha$-entmax function, ``$\eta$'' in the final loss function, the number of EWSGAT layers from $\{$0.0, 0.3, 0.5, 0.8, 1.0$\}$, $\{$1.2, 1.3, 1.4, 1.45, 1.5, 1.55, 1.6, 1.7, 1.8, 1.9$\}$, $\{$0.02, 0.025, 0.03, 0.035, 0.04, 0.06, 0.08$\}$, $\{$2, 3, 4, 5, 6$\}$, respectively. The Adam optimizer is employed with a learning rate of 0.005. 

\subsection{Clustering Performance Comparisons}
The first two datasets are noisy due to human interventions in their construction. In contrast, the third dataset has minimal human intervention. We introduce random noise edges into the low-intervention dataset in different proportions denoted by the variable ``N''. We randomly add 10$\%$ and 20$\%$ of edges to ML100k to create new experimental datasets. 

For all methods, we repeat each experiment 5 times by changing the random seeds and report the average results. Table~\ref{baseline} presents a comparison of the clustering performance between CeeGCN and five mainstream graph clustering methods across multiple datasets.

Overall, CeeGCN demonstrates superior clustering capability in all datasets. Notably, the performance of CeeGCN on the vessel datasets significantly surpasses that on ML100k. This is attributed to CeeGCN's effective implementation of a sparsity-aware method, which reduces computational complexity and noise interference in dense graphs, thereby improving clustering performance. Furthermore, CeeGCN consistently performs well on ML100k, showcasing its ability to accurately identify and handle noisy edges. In particular, introducing 10\% and 20\% noisy edges leads to varying degrees of performance degradation across all clustering methods, with the F1 showing a particularly marked decline. This indicates that noisy edges weaken the separability of inter-cluster boundaries. Except for SSGCN, the comparative methods assume the reliability of the graph structure, thereby underperforming our approach when considering both metrics. Although SSGCN adopts graph structure learning, it fails to jointly optimize node representations and clustering, thus performing not well. By fusing edge weights in attention coefficient computation and incorporating the $\alpha$-entmax function to suppress noisy connections, our model can focus on related neighbors, thereby performing well. 

\subsection{Ablation Study}
To prove the effectiveness of the key designs in CeeGCN, we consider different model variants from four perspectives:
\begin{itemize}
\item Effect of cluster-oriented graph contraction module. We replace cluster-oriented graph contraction with random sampling to select nodes of similar size, retaining all connecting edges between the selected nodes to obtain a subgraph. We refer to this variant as CeeGCN-CGC.
\begin{table}[]
\begin{center}
\setcounter{table}{2}
\footnotesize
{\caption{Ablation study of key designs in CeeGCN.}\label{xiaorong}}

\begin{tabular}{l|cc|cc|cc}
\toprule
\multirow{2}{*}{Model} & \multicolumn{2}{c|}{Vessel01} & \multicolumn{2}{c|}{Vessel10} & \multicolumn{2}{c}{ML100K} \\
\cmidrule(lr){2-7} 
& ACC           & F1           & ACC           & F1           & ACC          & F1          \\ 
\midrule
CeeGCN-CGC                 & 78.42         & 87.90        & 74.59         & 85.45        & 43.63        & 60.76       \\ 
CeeGCN-EWSGAT                    & 53.67         & 44.23        & 53.19         & 37.86        & 35.74        & 42.13       \\
CeeGCN-entmax                    & 75.79         & 81.08        & 70.90         & 76.29        & 44.85        & 61.30       \\
CeeGCN-$f_{iz}$                    & 69.68         & 72.39        & 73.65         & 81.25        & 44.72        & 61.54       \\ 
CeeGCN-EWO                    & 77.14         & 85.06        & 75.09         & 84.59        & 44.97        & 62.04       \\
CeeGCN w/o Contraction                    & 78.23         & 87.14        & \textbf{75.81}         & 84.94        & 44.96        & 61.27       \\  \midrule

CeeGCN                   & \textbf{78.97}         & \textbf{88.25}        & 75.36         & \textbf{85.94}        & \textbf{46.36}        & \textbf{63.35}       \\
\bottomrule
\end{tabular}
\end{center}
\end{table}

\begin{table}[]
\begin{center}
\small
{\caption{The training time (in seconds) of CeeGCN with and without the contraction module.}\label{per}}
\begin{tabular}{l|r|r}
\hline
\textbf{Dataset} & \makecell[c]{\textbf{CeeGCN w/o}\\ \textbf{Contraction}} & \textbf{CeeGCN} \\ \hline
Vessel01        & 11,846.10 & 2,576.40 \\
Vessel10        & 17,767.93 & 3,701.65 \\
ML100k          & 8,492.50  &  994.30 \\
\hline
\end{tabular}
\end{center}
\end{table}

\item Effect of EWSGAT. We replace EWSGAT with vanilla GAT and refer to this variant as CeeGCN-EWSGAT.
\item Effect of the $\alpha$-entmax function. We replace the $\alpha$-entmax function with the softmax function and refer to this variant as CeeGCN-entmax.
\item Effect of the impact factor derived from edge weights. We remove the $f_{iz}$ to observe the changes in clustering results and refer to this variant as CeeGCN-$f_{iz}$.
\item Effect of edge weight optimization. We utilize the original edge weights $w_{ij}$ to get the modularity loss. We refer to this variant as CeeGCN-EWO.
\item Effect of the graph contraction. We remove the graph contraction module to observe the changes in clustering performance. We refer to this variant as CeeGCN w/o Contraction.

\end{itemize}

Table~\ref{xiaorong} reports the results. CeeGCN-CGC performs worse than CeeGCN. This indicates that randomly sampling nodes fails to effectively preserve the original graph's clustering structure, whereas the cluster-oriented graph contraction module can achieve this. CeeGCN-EWSGAT performs far worse than CeeGCN. This is because, for graphs with noisy edges, vanilla GAT lacks the ability to block unrelated neighbors from participating in message propagation, leading to a substantial drop in clustering performance. CeeGCN-EWO performs worse than CeeGCN. This indicates that CeeGCN can better optimize edge weights based on feedback during the training process, while fixing edge weights restricts edge weight optimization. The performance degradation of CeeGCN-entmax is attributed to the non-zero weights assigned by the softmax function, which leads to noise accumulation. In contrast, $\alpha$-entmax truncates low-correlation weights to zero through an automatic sparsification mechanism. This effectively blocks noise propagation and focuses on key neighbors. The performance degradation of CeeGCN w/o Weight validates the importance of edge weights as prior structural information. Removing $f_{iz}$ leads to the loss of connection strength information, whereas retaining this factor enables EWSGAT to integrate feature similarity with topological strength to guide the clustering task.

By comparing CeeGCN w/o Contraction and CeeGCN, we find that CeeGCN performs better than CeeGCN w/o Contraction in most cases. This may be because the original graph contains more noisy edges, whereas the subgraph consists of important nodes and fewer noisy edges. Training based on the subgraph enables the model to more easily acquire the ability to identify noisy edges. 
Assuming E is the number of edges, d is the hidden dimension of a single attention head, and there are H total heads per layer, the time complexity of each vanilla GAT layer is $O(E \cdot d \cdot H)$; while the time complexity of each EWSGAT layer is $O(E \cdot (d+1) \cdot H)$. Compared to unweighted graphs, storing the graph structure of weighted graphs in the form of tuples leads to a 50\% increase in GPU memory usage during the model training process. Therefore, graph Contracting is highly necessary.

We further compare the training time of CeeGCN with and without the contraction module. As shown in Table~\ref{per}, we find that incorporating the contraction module significantly improves the training efficiency. Meanwhile, the contraction module effectively reduces memory usage. Taking the Vessel01 dataset as an example: The original graph includes 818 nodes and 144,898 edges. Storing the graph structure requires approximately 1.74 MB. The extracted subgraph includes $\sim$425 nodes and $\sim$57183 edges. Storing these edges requires approximately 0.69 MB.

\subsection{Hyperparameter Analysis}
We select four important hyperparameters to study, including ``$\varepsilon$'' in the calculation of comprehensive scores, ``$\alpha$'' in the ``$\alpha$-entmax'' function, ``$\eta$'' in the final loss function, and the calculation method of the self-loop weight.

\begin{figure}
  \centering
  \includegraphics[width=1\linewidth,height=0.3\linewidth]{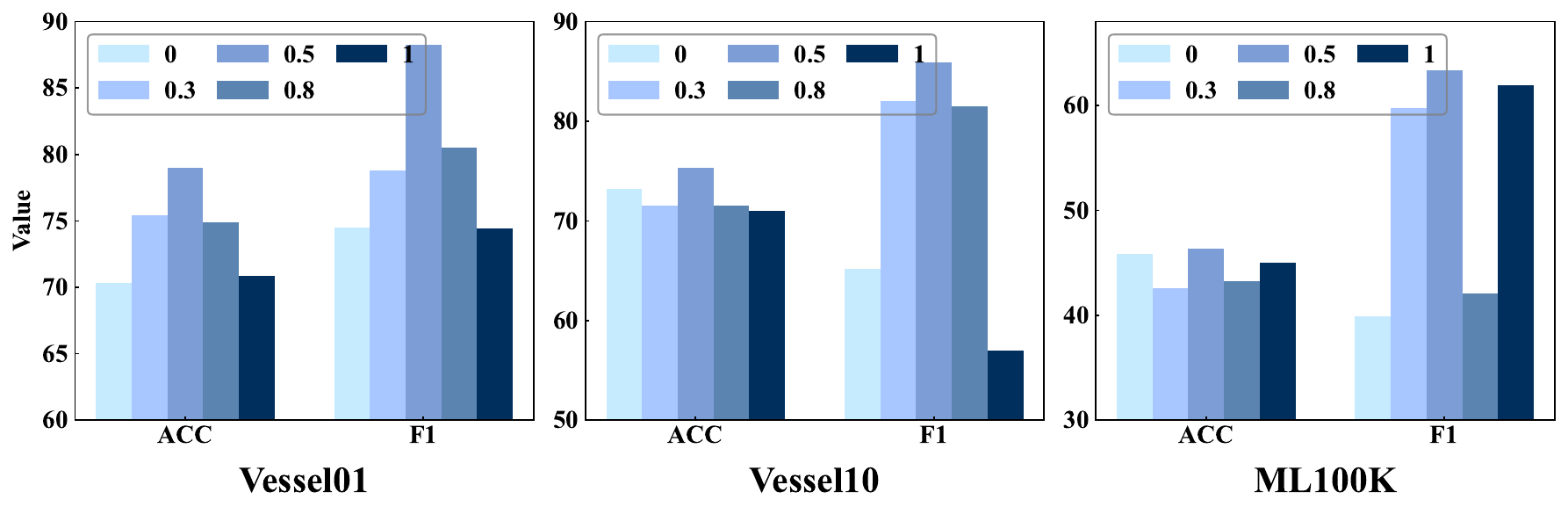}
  \vspace{-4mm}
  \caption{Analysis of $\varepsilon$ on model performance.}
  \vspace{-3mm}
  \label{hyperparameter}
\end{figure}

\begin{figure}[t]
    \centering
    \subfigure[Study of $\alpha$.]{
        \includegraphics[scale=0.2]{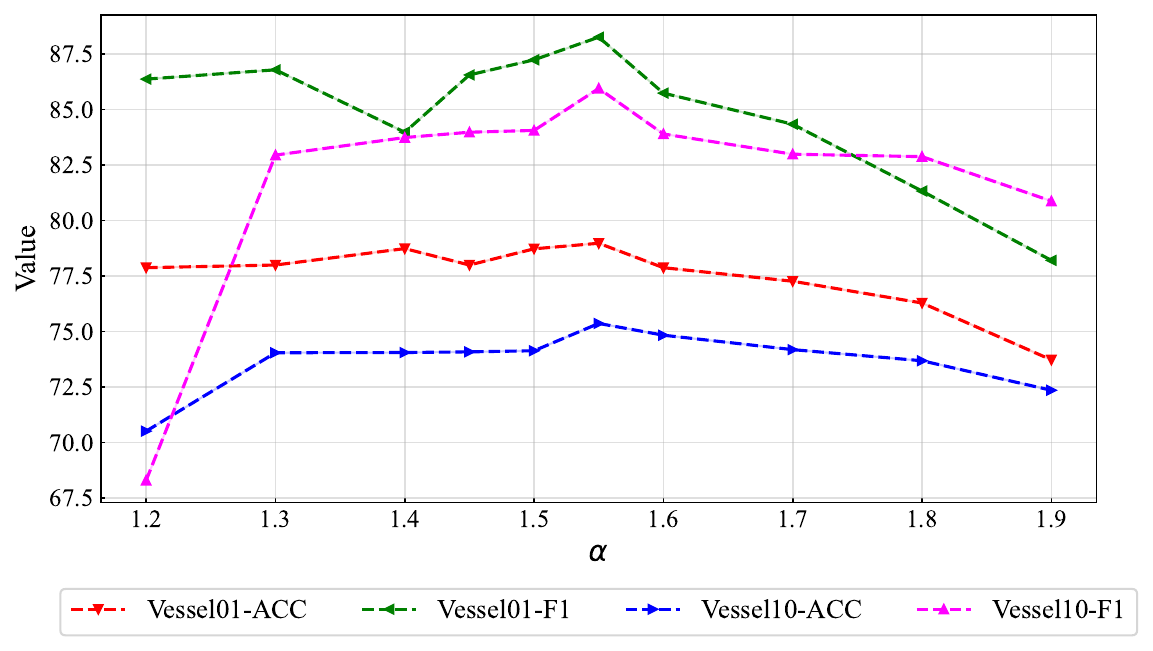}
        }
    \subfigure[Study of $\eta$]{
        \includegraphics[scale=0.2]{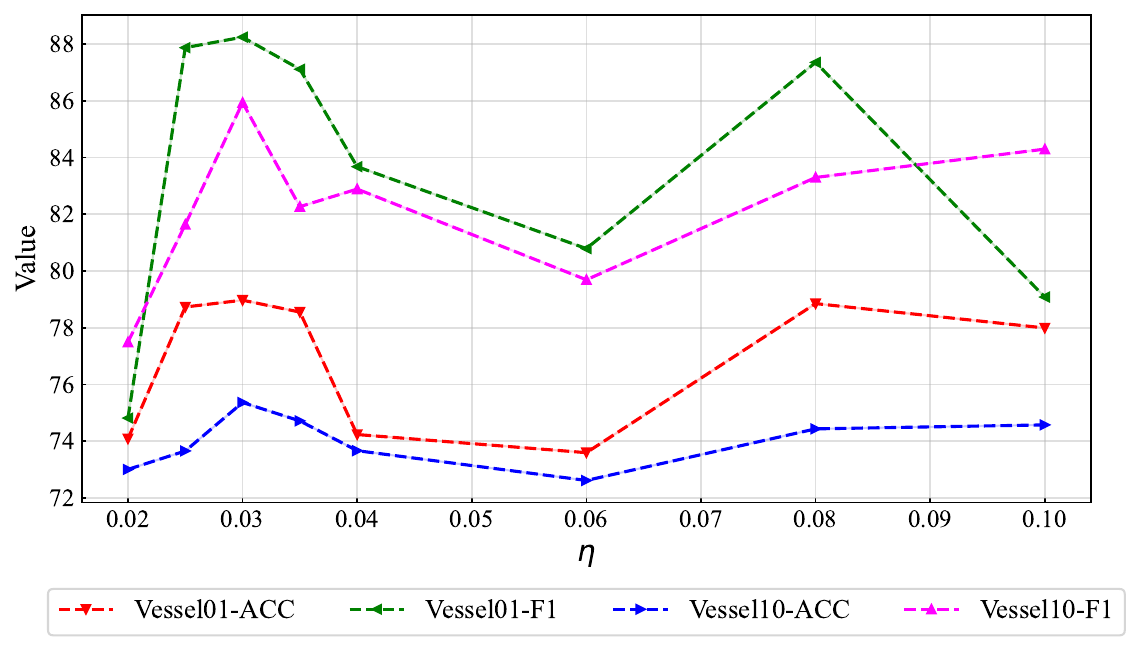}
    }
    \vspace{-3mm}
    \caption{Analysis of $\alpha$ and $\eta$ on model performance.}
    \vspace{-3mm}
    \label{threehyper}
\end{figure}

\begin{figure}
  \centering
  \includegraphics[width=0.8\linewidth]{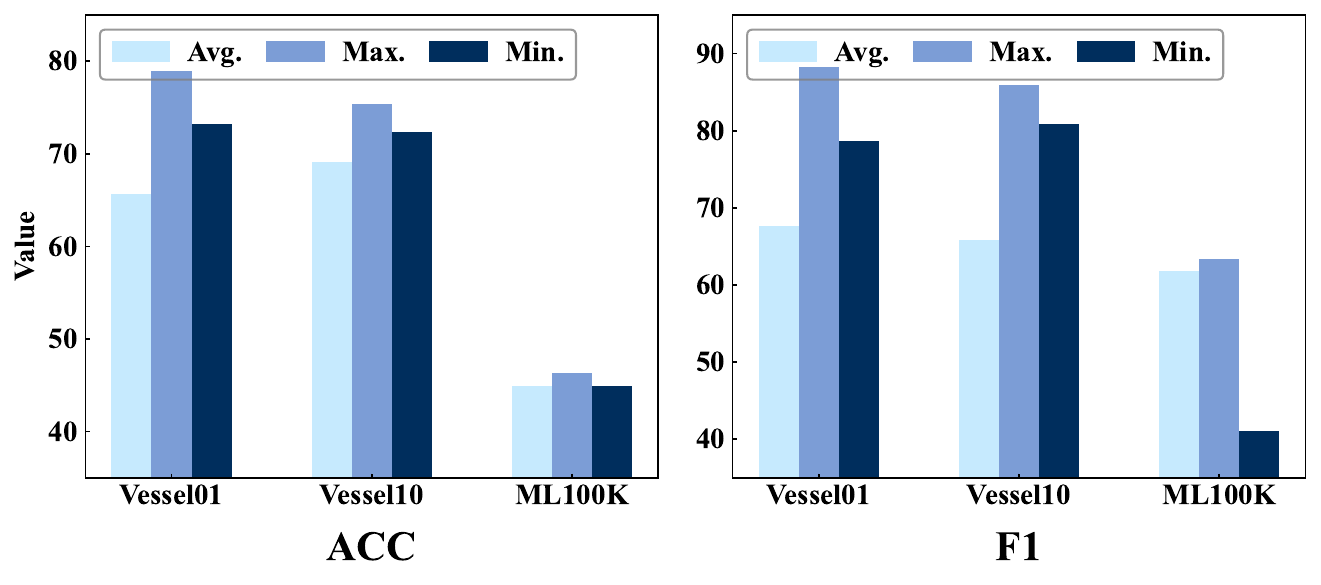}
  \vspace{-3mm}
  \caption{Clustering results under different self-loop weight calculation methods.}
  \vspace{-3mm}
  \label{self-loop}
\end{figure}

Figure~\ref{hyperparameter} presents the sensitivity analysis of the value of ``$\varepsilon$''. Experiment results indicate that the model achieves optimal performance when ``$\varepsilon$'' is set to 0.5. This phenomenon suggests that both node density and distance information are of equal importance. In Figure~\ref{threehyper}(a), optimal performance is achieved when ``$\alpha$'' is set to 1.55. When the ``$\alpha$'' $<$ 1.55, the $\alpha$-entmax function behaves more like softmax function, focusing on all neighbors, which may incorporate unrelated nodes during message passing. However, when the ``$\alpha$'' $>$ 1.55, it might overly suppress edge weight information, leading to performance degradation. In Figure~\ref{threehyper}(b), optimal performance is achieved when ``$\eta$'' is set to 0.03. This indicates the graph structure-based loss plays a more significant role.
As shown in Figure~\ref{self-loop}, we assign different weight strategies to the self-connected edges of the nodes. These strategies included the average, maximum and minimum values of the weights between a node and its neighbors. Employing the maximum value as the self-loop weight strategy generally produces the best performance across all datasets. By setting the self-weight to the maximum value, we strengthen the node's focus on its own information.

\subsection{Case Study}
To validate CeeGCN's robustness to noisy edges in graph clustering, we randomly select 10 nodes from the ML100K$_{N=10\%}$ dataset, each having 10 neighboring nodes.

We use CeeGCN w/o EWSGAT and CeeGCN to compute inter-node attention coefficients, and then visualize the results using heatmaps, as shown in Figure~\ref{hot}. Darker colors indicate a tighter relationship between nodes. The y-axis lists the node IDs, and the x-axis represents the neighbors of node, ordered by ID in ascending order.

In Figure~\ref{hot}(a), we observe that node 3 has strong attention connections with its 4th and 10th neighbors, indicating a close relationship. In Figure~\ref{hot}(b), we find that the attention coefficients between node 3 and three of its neighbors (nodes 1, 6, and 8) are set to 0 (highlighted by shaded blocks), identifying these edges as noisy edges. In contrast, the connections with the 4th and 10th neighbors show noticeably deeper colors, signifying their increased importance and relevance. This shows that CeeGCN effectively prioritizes neighbors with stronger relationships while ignoring noisy edges. 

To further analyze the effectiveness of noisy edge handling, we conduct a detailed comparison of the average attention coefficients assigned to noisy and noiseless edges. Since both edge weights and the number of neighbors influence attention coefficient computation, we filter the dataset to include 604 nodes with an average edge weight greater than 1.05. The 604 nodes are grouped by their number of neighbors: Group 1 (0 $<$ \#neighbors $\leq$ 25), Group 2 (26 $<$ \#neighbors $\leq$ 50), and Group 3 (51 $<$ \#neighbors $\leq$ 100). Table~\ref{case} presents the distribution of attention coefficients within these groups.
\begin{figure}[t]
\centering
\subfigure[{CeeGCN-EWSGAT}]{
\includegraphics[scale=0.25]{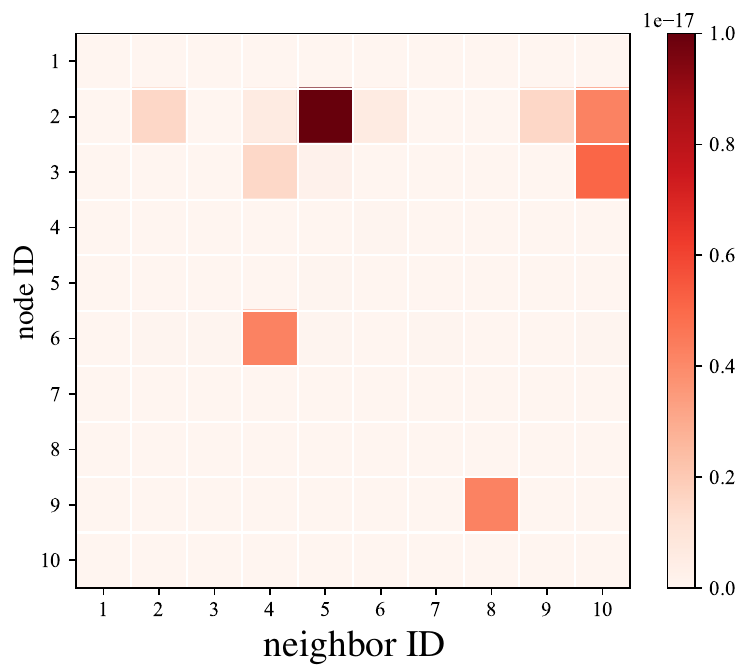}}
\subfigure[{CeeGCN}]{
\includegraphics[scale=0.25]{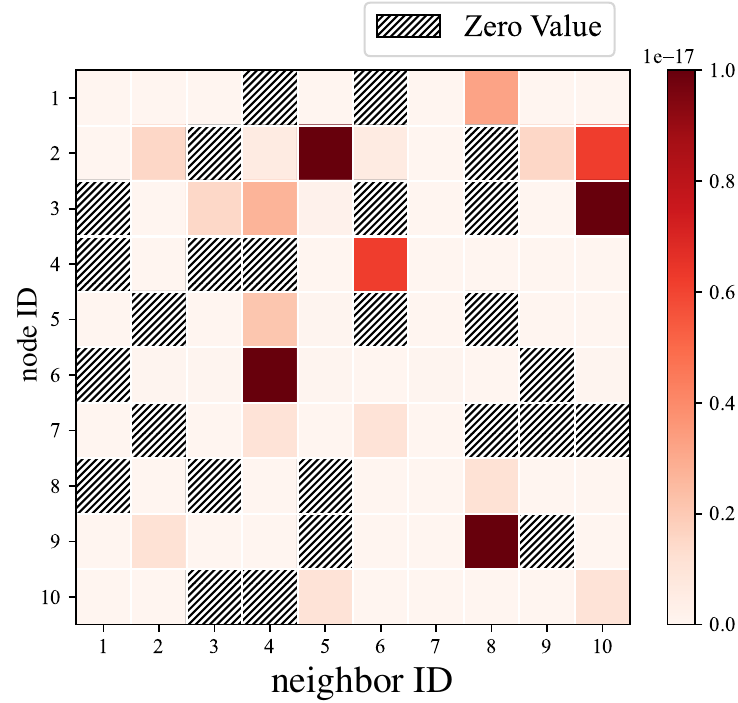}}
\vspace{-5mm}
\caption{Distributions of attention coefficients.}
\label{hot}
\end{figure}

Table~\ref{case} reveals that neighboring nodes connected by noisy edges consistently receive lower attention coefficients, especially when the number of a node's neighbors is small, indicating a more pronounced impact of noisy edges in such cases. As the number of neighbors increases, CeeGCN becomes more accurate at identifying important nodes, thereby better suppressing the influence of noisy edges. This demonstrates the effectiveness of combining edge weight information with node features, with model performance significantly improving as the number of neighbors grows. 

\begin{table}[t]
\vspace{-5mm}
\begin{center}
\small
\caption{Average normalized attention coefficients.}
\vspace{-3mm}
\label{case}
\begin{tabular}{l|c|c}
\toprule
\textbf{Group} & \textbf{Noiseless edge} & \textbf{Noisy edge} \\ \midrule
Group 1        & 0.050 & 0.015 \\
Group 2        & 0.029 & 0.007 \\
Group 3        & 0.024 & 0.005 \\
\bottomrule
\end{tabular}
\end{center}
\vspace{-5mm}
\end{table}

\section{Conclusion}
In this paper, we propose a weighted graph clustering model. We design a cluster-oriented graph contraction module. The module captures both intra-cluster and inter-cluster structural characteristics, significantly reducing the graph scale by sampling important nodes. We design an edge-weight-aware sparse graph attention network. The network calculates attention coefficients by considering both node features and edge weights. Noise edges are suppressed via the $\alpha$-entmax function. We will explore the scalability of our model on larger-scale weighted graphs in the future.

\section*{Acknowledgments}
This research is supported in part by the National Science Foundation of China (No. 62302469, No. 62176243), the Natural Science Foundation of Shandong Province (ZR2023QF100, ZR2022QF050, ZR2025QB43), and the Fundamental Research Funds for the Central Universities (202513026).

\balance

\bibliographystyle{ACM-Reference-Format}

\begin{thebibliography}{44}


\ifx \showCODEN    \undefined \def \showCODEN     #1{\unskip}     \fi
\ifx \showISBNx    \undefined \def \showISBNx     #1{\unskip}     \fi
\ifx \showISBNxiii \undefined \def \showISBNxiii  #1{\unskip}     \fi
\ifx \showISSN     \undefined \def \showISSN      #1{\unskip}     \fi
\ifx \showLCCN     \undefined \def \showLCCN      #1{\unskip}     \fi
\ifx \shownote     \undefined \def \shownote      #1{#1}          \fi
\ifx \showarticletitle \undefined \def \showarticletitle #1{#1}   \fi
\ifx \showURL      \undefined \def \showURL       {\relax}        \fi
\providecommand\bibfield[2]{#2}
\providecommand\bibinfo[2]{#2}
\providecommand\natexlab[1]{#1}
\providecommand\showeprint[2][]{arXiv:#2}

\bibitem[Ahmadi et~al\mbox{.}(2024)]%
        {ahmadi2024deep}
\bibfield{author}{\bibinfo{person}{Maedeh Ahmadi}, \bibinfo{person}{Mehran Safayani}, {and} \bibinfo{person}{Abdolreza Mirzaei}.} \bibinfo{year}{2024}\natexlab{}.
\newblock \showarticletitle{Deep graph clustering via mutual information maximization and mixture model}.
\newblock \bibinfo{journal}{\emph{KAIS}}  \bibinfo{volume}{66} (\bibinfo{year}{2024}), \bibinfo{pages}{4549--4572}.
\newblock


\bibitem[Balin and {\c{C}}ataly{\"u}rek(2023)]%
        {balin2023layer}
\bibfield{author}{\bibinfo{person}{Muhammed~Fatih Balin} {and} \bibinfo{person}{{\"U}mit {\c{C}}ataly{\"u}rek}.} \bibinfo{year}{2023}\natexlab{}.
\newblock \showarticletitle{Layer-Neighbor Sampling---Defusing Neighborhood Explosion in GNNs}. In \bibinfo{booktitle}{\emph{NeurIPS}}. \bibinfo{pages}{25819--25836}.
\newblock


\bibitem[Bezdek(1981)]%
        {bezdek1981objective}
\bibfield{author}{\bibinfo{person}{James~C Bezdek}.} \bibinfo{year}{1981}\natexlab{}.
\newblock \showarticletitle{Objective function clustering}.
\newblock In \bibinfo{booktitle}{\emph{Pattern recognition with fuzzy objective function algorithms}}. \bibinfo{pages}{43--93}.
\newblock


\bibitem[Bhowmick et~al\mbox{.}(2024)]%
        {bhowmick2024dgcluster}
\bibfield{author}{\bibinfo{person}{Aritra Bhowmick}, \bibinfo{person}{Mert Kosan}, \bibinfo{person}{Zexi Huang}, \bibinfo{person}{Ambuj Singh}, {and} \bibinfo{person}{Sourav Medya}.} \bibinfo{year}{2024}\natexlab{}.
\newblock \showarticletitle{DGCLUSTER: A neural framework for attributed graph clustering via modularity maximization}. In \bibinfo{booktitle}{\emph{AAAI}}. \bibinfo{pages}{11069--11077}.
\newblock


\bibitem[Blondel et~al\mbox{.}(2019)]%
        {ref18}
\bibfield{author}{\bibinfo{person}{Mathieu Blondel}, \bibinfo{person}{Andr{\'{e}} F.~T. Martins}, {and} \bibinfo{person}{Vlad Niculae}.} \bibinfo{year}{2019}\natexlab{}.
\newblock \showarticletitle{Learning classifiers with fenchel-young losses: Generalized entropies, margins, and algorithms}. In \bibinfo{booktitle}{\emph{AISTATS}}. \bibinfo{pages}{606--615}.
\newblock


\bibitem[Bo et~al\mbox{.}(2020)]%
        {DBLP:conf/www/Bo0SZL020}
\bibfield{author}{\bibinfo{person}{Deyu Bo}, \bibinfo{person}{Xiao Wang}, \bibinfo{person}{Chuan Shi}, \bibinfo{person}{Meiqi Zhu}, \bibinfo{person}{Emiao Lu}, {and} \bibinfo{person}{Peng Cui}.} \bibinfo{year}{2020}\natexlab{}.
\newblock \showarticletitle{Structural Deep Clustering Network}. In \bibinfo{booktitle}{\emph{WWW}}.
\newblock


\bibitem[Boutalbi et~al\mbox{.}(2024)]%
        {DBLP:conf/www/BoutalbiBVS24}
\bibfield{author}{\bibinfo{person}{Karima Boutalbi}, \bibinfo{person}{Rafika Boutalbi}, \bibinfo{person}{Herv{\'{e}} Verjus}, {and} \bibinfo{person}{Kav{\'{e}} Salamatian}.} \bibinfo{year}{2024}\natexlab{}.
\newblock \showarticletitle{Hierarchical Tensor Clustering for Multiple Graphs Representation}. In \bibinfo{booktitle}{\emph{WWW}}.
\newblock


\bibitem[Chunjing et~al\mbox{.}(2025)]%
        {chunjing2025learning}
\bibfield{author}{\bibinfo{person}{Xiao Chunjing}, \bibinfo{person}{Ranhao Guo}, \bibinfo{person}{Zhang Yongwang}, {and} \bibinfo{person}{Xiaoming Wu}.} \bibinfo{year}{2025}\natexlab{}.
\newblock \showarticletitle{Learning Multi-interest Embedding with Dynamic Graph Cluster for Sequention Recommendation}. In \bibinfo{booktitle}{\emph{UAI}}. \bibinfo{pages}{4652--4662}.
\newblock


\bibitem[Clevert et~al\mbox{.}(2016)]%
        {clevert2015fast}
\bibfield{author}{\bibinfo{person}{Djork{-}Arn{\'{e}} Clevert}, \bibinfo{person}{Thomas Unterthiner}, {and} \bibinfo{person}{Sepp Hochreiter}.} \bibinfo{year}{2016}\natexlab{}.
\newblock \showarticletitle{Fast and Accurate Deep Network Learning by Exponential Linear Units (ELUs)}. In \bibinfo{booktitle}{\emph{ICLR}}.
\newblock


\bibitem[Ding et~al\mbox{.}(2023)]%
        {ref28}
\bibfield{author}{\bibinfo{person}{Shifei Ding}, \bibinfo{person}{Benyu Wu}, \bibinfo{person}{Xiao Xu}, \bibinfo{person}{Lili Guo}, {and} \bibinfo{person}{Ling Ding}.} \bibinfo{year}{2023}\natexlab{}.
\newblock \showarticletitle{Graph clustering network with structure embedding enhanced}.
\newblock \bibinfo{journal}{\emph{PR}}  \bibinfo{volume}{144} (\bibinfo{year}{2023}).
\newblock


\bibitem[Gupta et~al\mbox{.}(2008)]%
        {gupta2008fast}
\bibfield{author}{\bibinfo{person}{Manish Gupta}, \bibinfo{person}{Amit Pathak}, {and} \bibinfo{person}{Soumen Chakrabarti}.} \bibinfo{year}{2008}\natexlab{}.
\newblock \showarticletitle{Fast algorithms for topk personalized pagerank queries}. In \bibinfo{booktitle}{\emph{ACM WWW}}. \bibinfo{pages}{1225--1226}.
\newblock


\bibitem[Hao and Zhu(2023)]%
        {hao2023deep}
\bibfield{author}{\bibinfo{person}{Jie Hao} {and} \bibinfo{person}{William Zhu}.} \bibinfo{year}{2023}\natexlab{}.
\newblock \showarticletitle{Deep graph clustering with enhanced feature representations for community detection}.
\newblock \bibinfo{journal}{\emph{Applied Intelligence}}  \bibinfo{volume}{53} (\bibinfo{year}{2023}), \bibinfo{pages}{1336--1349}.
\newblock


\bibitem[He et~al\mbox{.}(2021)]%
        {ref31}
\bibfield{author}{\bibinfo{person}{Chaobo He}, \bibinfo{person}{Yulong Zheng}, \bibinfo{person}{Xiang Fei}, \bibinfo{person}{Hanchao Li}, \bibinfo{person}{Zeng Hu}, {and} \bibinfo{person}{Yong Tang}.} \bibinfo{year}{2021}\natexlab{}.
\newblock \showarticletitle{Boosting nonnegative matrix factorization based community detection with graph attention auto-encoder}.
\newblock \bibinfo{journal}{\emph{TBD}}  \bibinfo{volume}{8} (\bibinfo{year}{2021}), \bibinfo{pages}{968--981}.
\newblock


\bibitem[He et~al\mbox{.}(2023)]%
        {he2023graph}
\bibfield{author}{\bibinfo{person}{Xiaxia He}, \bibinfo{person}{Boyue Wang}, \bibinfo{person}{Ruikun Li}, \bibinfo{person}{Junbin Gao}, \bibinfo{person}{Yongli Hu}, \bibinfo{person}{Guangyu Huo}, {and} \bibinfo{person}{Baocai Yin}.} \bibinfo{year}{2023}\natexlab{}.
\newblock \showarticletitle{Graph structure learning layer and its graph convolution clustering application}.
\newblock \bibinfo{journal}{\emph{Neural Networks}}  \bibinfo{volume}{165} (\bibinfo{year}{2023}), \bibinfo{pages}{1010--1020}.
\newblock


\bibitem[Huo et~al\mbox{.}(2021)]%
        {ref33}
\bibfield{author}{\bibinfo{person}{Guangyu Huo}, \bibinfo{person}{Yong Zhang}, \bibinfo{person}{Junbin Gao}, \bibinfo{person}{Boyue Wang}, \bibinfo{person}{Yongli Hu}, {and} \bibinfo{person}{Baocai Yin}.} \bibinfo{year}{2021}\natexlab{}.
\newblock \showarticletitle{CaEGCN: Cross-attention fusion based enhanced graph convolutional network for clustering}.
\newblock \bibinfo{journal}{\emph{TKDE}}  \bibinfo{volume}{35} (\bibinfo{year}{2021}), \bibinfo{pages}{3471--3483}.
\newblock


\bibitem[Huo et~al\mbox{.}(2023)]%
        {huo2023caegcn}
\bibfield{author}{\bibinfo{person}{Guangyu Huo}, \bibinfo{person}{Yong Zhang}, \bibinfo{person}{Junbin Gao}, \bibinfo{person}{Boyue Wang}, \bibinfo{person}{Yongli Hu}, {and} \bibinfo{person}{Baocai Yin}.} \bibinfo{year}{2023}\natexlab{}.
\newblock \showarticletitle{CaEGCN: Cross-Attention Fusion Based Enhanced Graph Convolutional Network for Clustering}.
\newblock \bibinfo{journal}{\emph{TKDE}}  \bibinfo{volume}{35} (\bibinfo{year}{2023}), \bibinfo{pages}{3471--3483}.
\newblock


\bibitem[Jin et~al\mbox{.}(2019)]%
        {jin2019graph}
\bibfield{author}{\bibinfo{person}{Di Jin}, \bibinfo{person}{Ziyang Liu}, \bibinfo{person}{Weihao Li}, \bibinfo{person}{Dongxiao He}, {and} \bibinfo{person}{Weixiong Zhang}.} \bibinfo{year}{2019}\natexlab{}.
\newblock \showarticletitle{Graph convolutional networks meet markov random fields: Semi-supervised community detection in attribute networks}. In \bibinfo{booktitle}{\emph{AAAI}}. \bibinfo{pages}{152--159}.
\newblock


\bibitem[Kang et~al\mbox{.}(2025)]%
        {kang2024cdc}
\bibfield{author}{\bibinfo{person}{Zhao Kang}, \bibinfo{person}{Xuanting Xie}, \bibinfo{person}{Bingheng Li}, {and} \bibinfo{person}{Erlin Pan}.} \bibinfo{year}{2025}\natexlab{}.
\newblock \showarticletitle{{CDC:} {A} Simple Framework for Complex Data Clustering}.
\newblock \bibinfo{journal}{\emph{IEEE Transactions on Neural Networks and Learning Systems}}  \bibinfo{volume}{36} (\bibinfo{year}{2025}), \bibinfo{pages}{13177--13188}.
\newblock


\bibitem[Kipf and Welling(2017)]%
        {DBLP:conf/iclr/KipfW17}
\bibfield{author}{\bibinfo{person}{Thomas~N. Kipf} {and} \bibinfo{person}{Max Welling}.} \bibinfo{year}{2017}\natexlab{}.
\newblock \showarticletitle{Semi-Supervised Classification with Graph Convolutional Networks}. In \bibinfo{booktitle}{\emph{ICLR}}.
\newblock


\bibitem[Kulatilleke et~al\mbox{.}(2025)]%
        {kulatilleke2025scgc}
\bibfield{author}{\bibinfo{person}{Gayan~K Kulatilleke}, \bibinfo{person}{Marius Portmann}, {and} \bibinfo{person}{Shekhar~S Chandra}.} \bibinfo{year}{2025}\natexlab{}.
\newblock \showarticletitle{SCGC: Self-supervised contrastive graph clustering}.
\newblock \bibinfo{journal}{\emph{Neurocomputing}}  \bibinfo{volume}{611} (\bibinfo{year}{2025}), \bibinfo{pages}{128629}.
\newblock


\bibitem[Liu et~al\mbox{.}(2024)]%
        {liu2023deep}
\bibfield{author}{\bibinfo{person}{Meng Liu}, \bibinfo{person}{Yue Liu}, \bibinfo{person}{Ke Liang}, \bibinfo{person}{Wenxuan Tu}, \bibinfo{person}{Siwei Wang}, \bibinfo{person}{Sihang Zhou}, {and} \bibinfo{person}{Xinwang Liu}.} \bibinfo{year}{2024}\natexlab{}.
\newblock \showarticletitle{Deep Temporal Graph Clustering}. In \bibinfo{booktitle}{\emph{ICLR}}.
\newblock


\bibitem[Liu et~al\mbox{.}(2023)]%
        {liu2023hard}
\bibfield{author}{\bibinfo{person}{Yue Liu}, \bibinfo{person}{Xihong Yang}, \bibinfo{person}{Sihang Zhou}, \bibinfo{person}{Xinwang Liu}, \bibinfo{person}{Zhen Wang}, \bibinfo{person}{Ke Liang}, \bibinfo{person}{Wenxuan Tu}, \bibinfo{person}{Liang Li}, \bibinfo{person}{Jingcan Duan}, {and} \bibinfo{person}{Cancan Chen}.} \bibinfo{year}{2023}\natexlab{}.
\newblock \showarticletitle{Hard sample aware network for contrastive deep graph clustering}. In \bibinfo{booktitle}{\emph{AAAI}}. \bibinfo{pages}{8914--8922}.
\newblock


\bibitem[Liu et~al\mbox{.}(2022)]%
        {liu2022towards}
\bibfield{author}{\bibinfo{person}{Yixin Liu}, \bibinfo{person}{Yu Zheng}, \bibinfo{person}{Daokun Zhang}, \bibinfo{person}{Hongxu Chen}, \bibinfo{person}{Hao Peng}, {and} \bibinfo{person}{Shirui Pan}.} \bibinfo{year}{2022}\natexlab{}.
\newblock \showarticletitle{Towards unsupervised deep graph structure learning}. In \bibinfo{booktitle}{\emph{ACM WWW}}. \bibinfo{pages}{1392--1403}.
\newblock


\bibitem[Pan and Kang(2023)]%
        {DBLP:conf/icml/Pan023}
\bibfield{author}{\bibinfo{person}{Erlin Pan} {and} \bibinfo{person}{Zhao Kang}.} \bibinfo{year}{2023}\natexlab{}.
\newblock \showarticletitle{Beyond Homophily: Reconstructing Structure for Graph-agnostic Clustering}. In \bibinfo{booktitle}{\emph{ICML}}.
\newblock


\bibitem[Peng et~al\mbox{.}(2023)]%
        {ref29}
\bibfield{author}{\bibinfo{person}{Zhihao Peng}, \bibinfo{person}{Hui Liu}, \bibinfo{person}{Yuheng Jia}, {and} \bibinfo{person}{Junhui Hou}.} \bibinfo{year}{2023}\natexlab{}.
\newblock \showarticletitle{Deep attention-guided graph clustering with dual self-supervision}.
\newblock \bibinfo{journal}{\emph{IEEE Transactions on Circuits and Systems for Video Technology}}  \bibinfo{volume}{33} (\bibinfo{year}{2023}), \bibinfo{pages}{3296--3307}.
\newblock


\bibitem[Peters et~al\mbox{.}(2019)]%
        {peters2019sparse}
\bibfield{author}{\bibinfo{person}{Ben Peters}, \bibinfo{person}{Vlad Niculae}, {and} \bibinfo{person}{Andr{\'{e}} F.~T. Martins}.} \bibinfo{year}{2019}\natexlab{}.
\newblock \showarticletitle{Sparse Sequence-to-Sequence Models}. In \bibinfo{booktitle}{\emph{ACL}}. \bibinfo{pages}{1504--1519}.
\newblock


\bibitem[Qiu et~al\mbox{.}(2022)]%
        {ref30}
\bibfield{author}{\bibinfo{person}{Chenyang Qiu}, \bibinfo{person}{Zhaoci Huang}, \bibinfo{person}{Wenzhe Xu}, {and} \bibinfo{person}{Huijia Li}.} \bibinfo{year}{2022}\natexlab{}.
\newblock \showarticletitle{VGAER: graph neural network reconstruction based community detection}.
\newblock \bibinfo{journal}{\emph{Association for the Advancement of Artificial Intelligence,}} (\bibinfo{year}{2022}).
\newblock


\bibitem[Shen and Kang(2025)]%
        {shen2025heterophily}
\bibfield{author}{\bibinfo{person}{Zhixiang Shen} {and} \bibinfo{person}{Zhao Kang}.} \bibinfo{year}{2025}\natexlab{}.
\newblock \showarticletitle{When Heterophily Meets Heterogeneous Graphs: Latent Graphs Guided Unsupervised Representation Learning}.
\newblock \bibinfo{journal}{\emph{IEEE Transactions on Neural Networks and Learning Systems}}  \bibinfo{volume}{36} (\bibinfo{year}{2025}), \bibinfo{pages}{10283--10296}.
\newblock


\bibitem[Shin et~al\mbox{.}(2023)]%
        {shin2023efficient}
\bibfield{author}{\bibinfo{person}{Seiyun Shin}, \bibinfo{person}{Ilan Shomorony}, {and} \bibinfo{person}{Han Zhao}.} \bibinfo{year}{2023}\natexlab{}.
\newblock \showarticletitle{Efficient learning of linear graph neural networks via node subsampling}. In \bibinfo{booktitle}{\emph{NeurIPS}}. \bibinfo{pages}{55479--55501}.
\newblock


\bibitem[Srichandra and Bhadra(2024)]%
        {srichandra2024community}
\bibfield{author}{\bibinfo{person}{Ippatapu~Venkata Srichandra} {and} \bibinfo{person}{Pratiti Bhadra}.} \bibinfo{year}{2024}\natexlab{}.
\newblock \showarticletitle{Community Detection Using Graph Attention Networks Clustering Algorithm}. In \bibinfo{booktitle}{\emph{IEEE International Conference for Convergence in Technology}}. \bibinfo{pages}{1--6}.
\newblock


\bibitem[Sun et~al\mbox{.}(2022)]%
        {sun2022graph}
\bibfield{author}{\bibinfo{person}{Qingyun Sun}, \bibinfo{person}{Jianxin Li}, \bibinfo{person}{Hao Peng}, \bibinfo{person}{Jia Wu}, \bibinfo{person}{Xingcheng Fu}, \bibinfo{person}{Cheng Ji}, {and} \bibinfo{person}{S~Yu Philip}.} \bibinfo{year}{2022}\natexlab{}.
\newblock \showarticletitle{Graph structure learning with variational information bottleneck}. In \bibinfo{booktitle}{\emph{AAAI}}. \bibinfo{pages}{4165--4174}.
\newblock


\bibitem[Velickovic et~al\mbox{.}(2018)]%
        {velickovic2017graph}
\bibfield{author}{\bibinfo{person}{Petar Velickovic}, \bibinfo{person}{Guillem Cucurull}, \bibinfo{person}{Arantxa Casanova}, \bibinfo{person}{Adriana Romero}, \bibinfo{person}{Pietro Li{\`{o}}}, {and} \bibinfo{person}{Yoshua Bengio}.} \bibinfo{year}{2018}\natexlab{}.
\newblock \showarticletitle{Graph Attention Networks}. In \bibinfo{booktitle}{\emph{ICLR}}.
\newblock


\bibitem[Wang et~al\mbox{.}(2022)]%
        {wang2022deep}
\bibfield{author}{\bibinfo{person}{Chun Wang}, \bibinfo{person}{Shirui Pan}, \bibinfo{person}{P~Yu Celina}, \bibinfo{person}{Ruiqi Hu}, \bibinfo{person}{Guodong Long}, {and} \bibinfo{person}{Chengqi Zhang}.} \bibinfo{year}{2022}\natexlab{}.
\newblock \showarticletitle{Deep neighbor-aware embedding for node clustering in attributed graphs}.
\newblock \bibinfo{journal}{\emph{PR}}  \bibinfo{volume}{122} (\bibinfo{year}{2022}), \bibinfo{pages}{108230}.
\newblock


\bibitem[Wang et~al\mbox{.}(2021)]%
        {wang2021graph}
\bibfield{author}{\bibinfo{person}{Ruijia Wang}, \bibinfo{person}{Shuai Mou}, \bibinfo{person}{Xiao Wang}, \bibinfo{person}{Wanpeng Xiao}, \bibinfo{person}{Qi Ju}, \bibinfo{person}{Chuan Shi}, {and} \bibinfo{person}{Xing Xie}.} \bibinfo{year}{2021}\natexlab{}.
\newblock \showarticletitle{Graph structure estimation neural networks}. In \bibinfo{booktitle}{\emph{ACM MM}}. \bibinfo{pages}{342--353}.
\newblock


\bibitem[Wang et~al\mbox{.}(2023)]%
        {wang2022ncagc}
\bibfield{author}{\bibinfo{person}{Tong Wang}, \bibinfo{person}{Junhua Wu}, \bibinfo{person}{Yaolei Qi}, \bibinfo{person}{Xiaoming Qi}, \bibinfo{person}{Juwei Guan}, \bibinfo{person}{Yuan Zhang}, {and} \bibinfo{person}{Guanyu Yang}.} \bibinfo{year}{2023}\natexlab{}.
\newblock \showarticletitle{Neighborhood contrastive representation learning for attributed graph clustering}.
\newblock \bibinfo{journal}{\emph{Neurocomputing}}  \bibinfo{volume}{562} (\bibinfo{year}{2023}), \bibinfo{pages}{126880}.
\newblock


\bibitem[Wen et~al\mbox{.}(2023)]%
        {wen2023efficient}
\bibfield{author}{\bibinfo{person}{Yi Wen}, \bibinfo{person}{Suyuan Liu}, \bibinfo{person}{Xinhang Wan}, \bibinfo{person}{Siwei Wang}, \bibinfo{person}{Ke Liang}, \bibinfo{person}{Xinwang Liu}, \bibinfo{person}{Xihong Yang}, {and} \bibinfo{person}{Pei Zhang}.} \bibinfo{year}{2023}\natexlab{}.
\newblock \showarticletitle{Efficient multi-view graph clustering with local and global structure preservation}. In \bibinfo{booktitle}{\emph{ACM MM}}. \bibinfo{pages}{3021--3030}.
\newblock


\bibitem[Yang et~al\mbox{.}(2022)]%
        {yang2022contrastive}
\bibfield{author}{\bibinfo{person}{Xihong Yang}, \bibinfo{person}{Yue Liu}, \bibinfo{person}{Sihang Zhou}, \bibinfo{person}{Siwei Wang}, \bibinfo{person}{Xinwang Liu}, {and} \bibinfo{person}{En Zhu}.} \bibinfo{year}{2022}\natexlab{}.
\newblock \showarticletitle{Contrastive deep graph clustering with learnable augmentation}.
\newblock \bibinfo{journal}{\emph{arXiv preprint arXiv:2212.03559}}  \bibinfo{volume}{abs/2212.03559} (\bibinfo{year}{2022}).
\newblock


\bibitem[You et~al\mbox{.}(2020)]%
        {you2020handling}
\bibfield{author}{\bibinfo{person}{Jiaxuan You}, \bibinfo{person}{Xiaobai Ma}, \bibinfo{person}{Daisy~Yi Ding}, \bibinfo{person}{Mykel~J. Kochenderfer}, {and} \bibinfo{person}{Jure Leskovec}.} \bibinfo{year}{2020}\natexlab{}.
\newblock \showarticletitle{Handling missing data with graph representation learning}. In \bibinfo{booktitle}{\emph{NeurIPS}}. \bibinfo{pages}{19075--19087}.
\newblock


\bibitem[Zeng et~al\mbox{.}(2020)]%
        {zeng2019graphsaint}
\bibfield{author}{\bibinfo{person}{Hanqing Zeng}, \bibinfo{person}{Hongkuan Zhou}, \bibinfo{person}{Ajitesh Srivastava}, \bibinfo{person}{Rajgopal Kannan}, {and} \bibinfo{person}{Viktor~K. Prasanna}.} \bibinfo{year}{2020}\natexlab{}.
\newblock \showarticletitle{Graphsaint: Graph sampling based inductive learning method}. In \bibinfo{booktitle}{\emph{ICLR}}.
\newblock


\bibitem[Zhang et~al\mbox{.}(2023)]%
        {zhang2023subgraph}
\bibfield{author}{\bibinfo{person}{Qi Zhang}, \bibinfo{person}{Yanfeng Sun}, \bibinfo{person}{Yongli Hu}, \bibinfo{person}{Shaofan Wang}, {and} \bibinfo{person}{Baocai Yin}.} \bibinfo{year}{2023}\natexlab{}.
\newblock \showarticletitle{A subgraph sampling method for training large-scale graph convolutional network}.
\newblock \bibinfo{journal}{\emph{Information Sciences}}  \bibinfo{volume}{649} (\bibinfo{year}{2023}), \bibinfo{pages}{119661}.
\newblock


\bibitem[Zhang et~al\mbox{.}(2022)]%
        {zhang2022hierarchical}
\bibfield{author}{\bibinfo{person}{Zaixi Zhang}, \bibinfo{person}{Qi Liu}, \bibinfo{person}{Qingyong Hu}, {and} \bibinfo{person}{Chee-Kong Lee}.} \bibinfo{year}{2022}\natexlab{}.
\newblock \showarticletitle{Hierarchical graph transformer with adaptive node sampling}. In \bibinfo{booktitle}{\emph{NeurIPS}}.
\newblock


\bibitem[Zhong et~al\mbox{.}(2021)]%
        {zhong2021graph}
\bibfield{author}{\bibinfo{person}{Huasong Zhong}, \bibinfo{person}{Jianlong Wu}, \bibinfo{person}{Chong Chen}, \bibinfo{person}{Jianqiang Huang}, \bibinfo{person}{Minghua Deng}, \bibinfo{person}{Liqiang Nie}, \bibinfo{person}{Zhouchen Lin}, {and} \bibinfo{person}{Xian{-}Sheng Hua}.} \bibinfo{year}{2021}\natexlab{}.
\newblock \showarticletitle{Graph contrastive clustering}. In \bibinfo{booktitle}{\emph{ICCV}}. \bibinfo{pages}{9224--9233}.
\newblock


\bibitem[Zhu et~al\mbox{.}(2024)]%
        {zhu2024every}
\bibfield{author}{\bibinfo{person}{Pengfei Zhu}, \bibinfo{person}{Qian Wang}, \bibinfo{person}{Yu Wang}, \bibinfo{person}{Jialu Li}, {and} \bibinfo{person}{Qinghua Hu}.} \bibinfo{year}{2024}\natexlab{}.
\newblock \showarticletitle{Every node is different: Dynamically fusing self-supervised tasks for attributed graph clustering}. In \bibinfo{booktitle}{\emph{AAAI}}. \bibinfo{pages}{17184--17192}.
\newblock


\bibitem[Zou et~al\mbox{.}(2019)]%
        {zou2019layer}
\bibfield{author}{\bibinfo{person}{Difan Zou}, \bibinfo{person}{Ziniu Hu}, \bibinfo{person}{Yewen Wang}, \bibinfo{person}{Song Jiang}, \bibinfo{person}{Yizhou Sun}, {and} \bibinfo{person}{Quanquan Gu}.} \bibinfo{year}{2019}\natexlab{}.
\newblock \showarticletitle{Layer-dependent importance sampling for training deep and large graph convolutional networks}. In \bibinfo{booktitle}{\emph{NeurIPS}}. \bibinfo{pages}{11247--11256}.
\newblock


\end{thebibliography}

\clearpage 
\appendix

\section{Appendix}
\subsection{Detailed Calculation of Threshold $\tau$}
We provide the detailed calculation process for the threshold $\tau$ in Algorithm~\ref{algorithm2}, utilizing the Bisection method to calculate the threshold with high efficiency.

\setcounter{algocf}{0}

\begin{algorithm}[b]
    \caption{Calculation of $\tau$ by Bisection}
    \label{algorithm2}
    \SetKwInOut{Input}{Input}
    \SetKwInOut{Output}{Output}
    \SetKwInOut{Define}{Define}
    \Input{$\bm{e_{i}^{\prime}}$: the redefined attention coefficients between node $i$ and its neighbors; $d$: dimension of the vector $\bm{e_{i}^{\prime}}$; $\alpha$: the hyperparameter of the $\alpha$-entmax function. }
    \Output{$\tau$: threshold.}
    \Define{
        ${p} = [(\alpha-1){e_{iz}^{\prime}}-\tau({e_{iz}^{\prime}})]_+^{\frac{1}{\alpha - 1}}$\;
        $\tau_{\min} = \max({\bm{e_{i}^{\prime}}}) - 1$\;
        $\tau_{\max} = \max({\bm{e_{i}^{\prime}}}) - d^{1-\alpha}$\;
    }
    \While{$\mathrm{sum} \neq 1$}{
        $\tau = (\tau_{\min} + \tau_{\max}) / 2$\;
        $\mathrm{sum} = \sum_{i=1}^{d} p_i(\tau)$\;
        \eIf{$\mathrm{sum} < 1$}{
            $\tau_{\max} = \tau$\;
        }{
            $\tau_{\min} = \tau$\;
        }
    }
    \Return $\tau$\;
\end{algorithm}

\subsection{Details of the Baselines}
We compare CeeGCN with several advanced graph clustering methods. The detailed introduction of these methods is as follows:
\begin{itemize}
\item \textbf{SSGCN}~\cite{he2023graph}\textbf{:} The method uses a graph structure learning layer and an adaptive graph convolutional layer to jointly optimize graph structure and node representations. By employing sparse representation, it effectively handles data interference, thereby enhancing clustering performance.
\item \textbf{DyFSS}~\cite{zhu2024every}\textbf{:} The method assigns each self-supervised learning (SSL) task to a task-specific GCN layer to extract features using the corresponding SSL loss.
\item \textbf{DGCluster}~\cite{bhowmick2024dgcluster}\textbf{:} The method proposes a trainable modular method that models the graph structure using GNN (e.g., GCN, GAT) to learn node representations, thereby accomplishing cluster number estimation.
\item \textbf{SCGC}~\cite{kulatilleke2025scgc}\textbf{:} The method based on graph autoencoders, aims to improve the efficiency and scalability of existing deep graph clustering methods by decoupling transformation and propagation operations.
\item \textbf{DAGC}~\cite{ref29}\textbf{:} This method employs various fusion strategies to integrate feature learning from both GCN and autoencoders, along with clustering distribution information, dynamically weighting their significance using attention mechanisms.
\item \textbf{GC-SEE}~\cite{ref28}\textbf{:} This model utilizes autoencoders for attribute feature extraction, graph attention autoencoders to learn node importance and structural information, adaptive convolution for fusing attribute and structural information.
\item \textbf{VGAER}~\cite{ref30}\textbf{:} The technique of graph variational self-encoders enables the reconstruction of non-linear modularity by integrating modularity information and graph structure.
\item \textbf{SUBLIME}~\cite{liu2022towards}\textbf{:} The method effectively learns the graph structure information by combining the learned adjacency matrix with data features and maximizing view consistency through node-level contrastive learning.
\item \textbf{NMFGAAE}~\cite{ref31}\textbf{:} The graph attention auto-encoder is utilized to learn the representation of nodes, and the cluster structure is revealed through a non-negative matrix decomposition of the node representation.
\item \textbf{CaEGCN}~\cite{ref33}\textbf{:} This model uses an attention mechanism to combine the distinct representations learned by the content self-encoder and graph self-encoder modules, which capture heterogeneous features.
\item \textbf{GMIM}~\cite{ahmadi2024deep}\textbf{:} The method proposes a contrastive learning framework that aims to directly optimize node embeddings beneficial for clustering. It introduces a gaussian mixture distribution in the embedding space, thereby promoting the generation of node representations suitable for the clustering task. GMIM also incorporates a graph diffusion mechanism to capture the global structural information of the graph.
\end{itemize}

\begin{figure}[b]
    \setcounter{figure}{0}
    \centering
    \subfigure[Ground Truth]{
        \includegraphics[width=0.3\linewidth,height=0.3\linewidth]{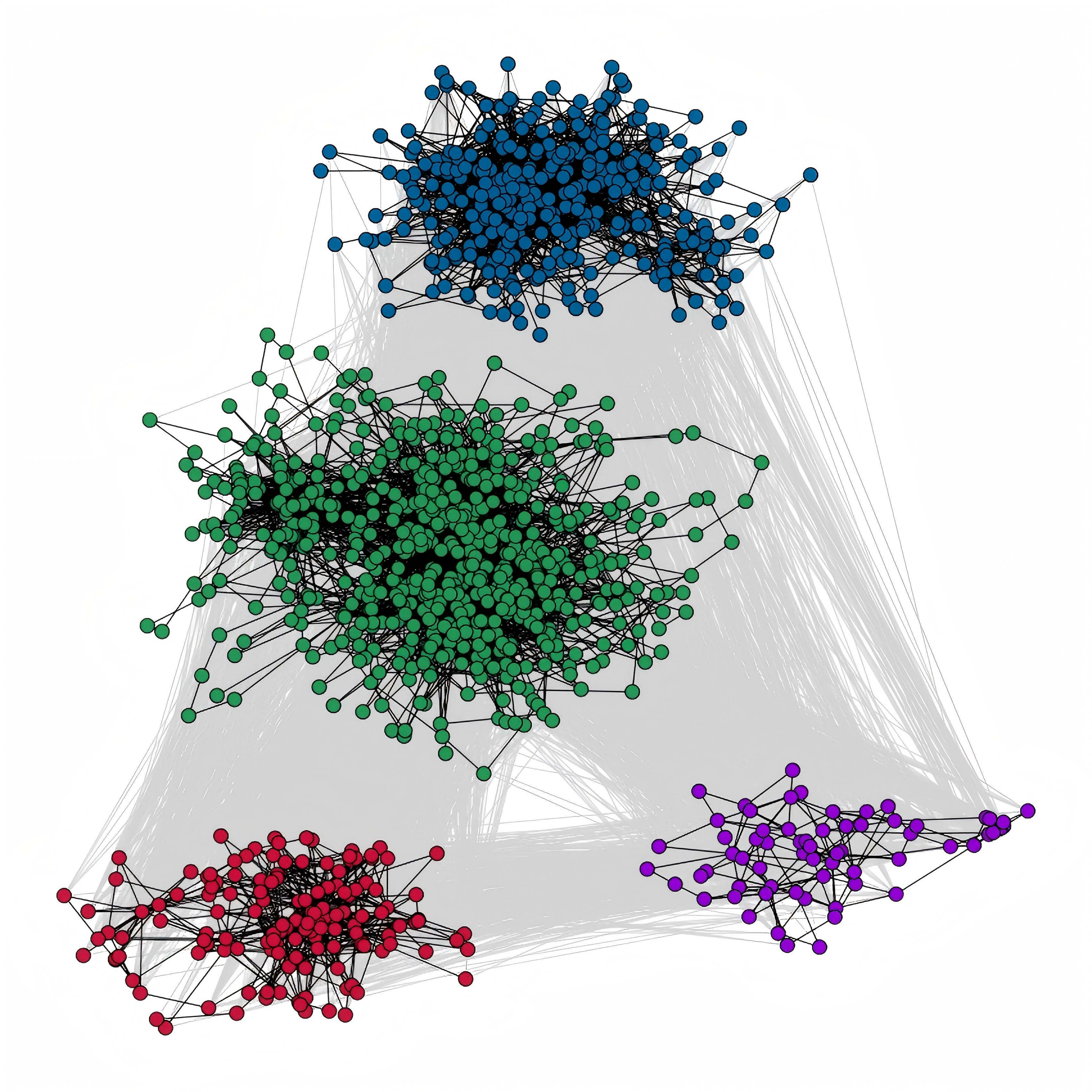}
    }
    \subfigure[SSGCN]{
        \includegraphics[width=0.3\linewidth,height=0.3\linewidth]{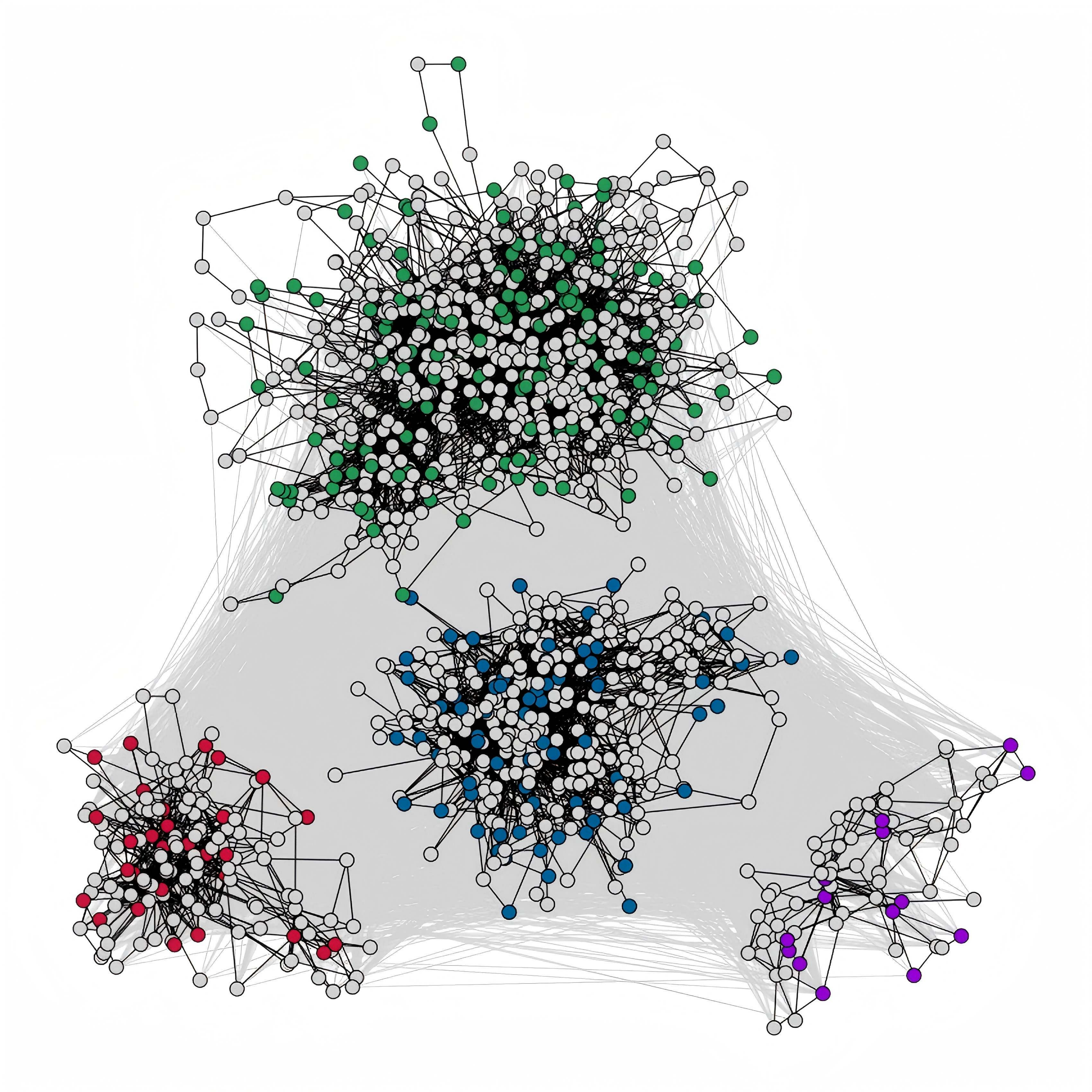}
    }
    \subfigure[DGCLUSTER]{
        \includegraphics[width=0.3\linewidth,height=0.3\linewidth]{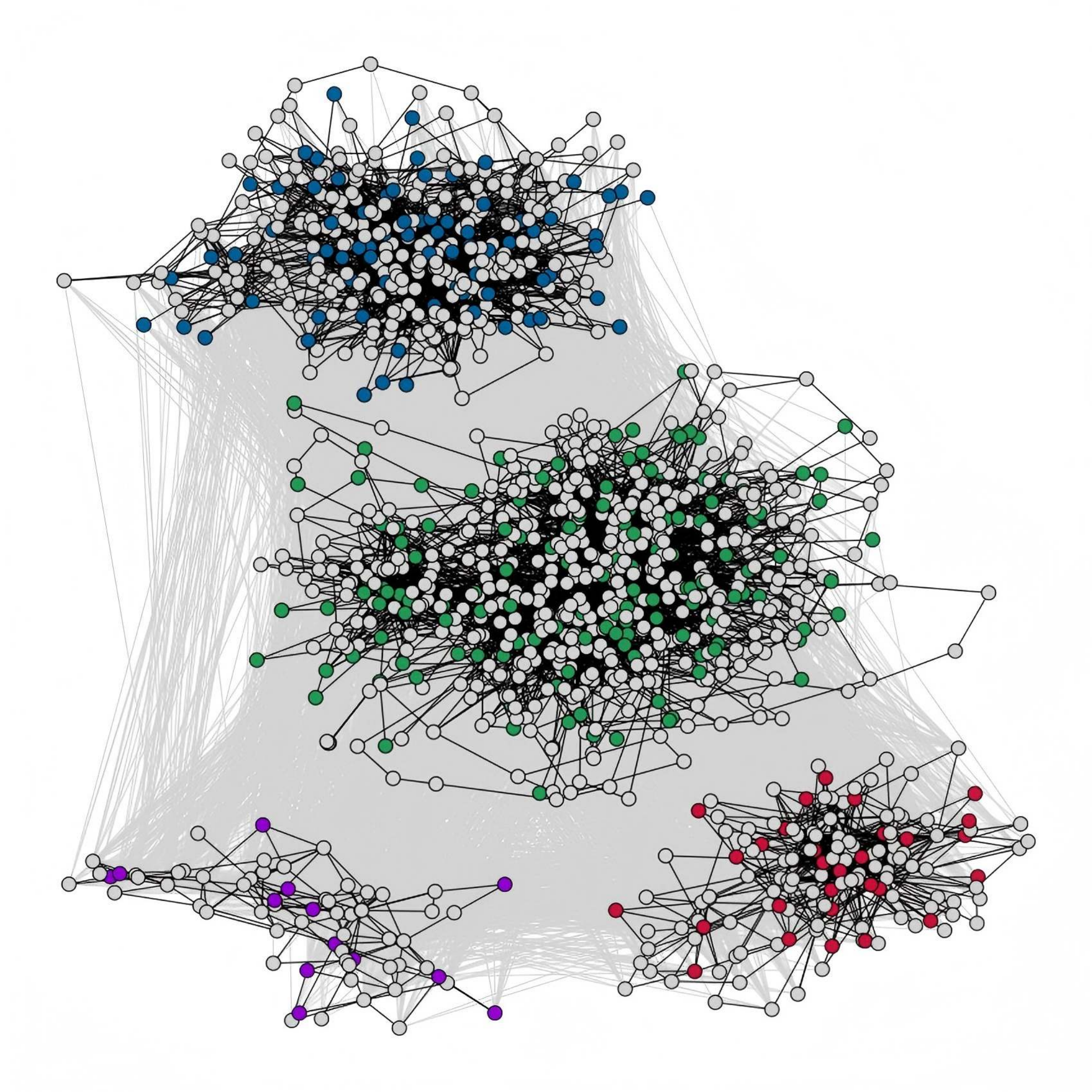}
    }
     \subfigure[SCGC]{
        \includegraphics[width=0.3\linewidth,height=0.3\linewidth]{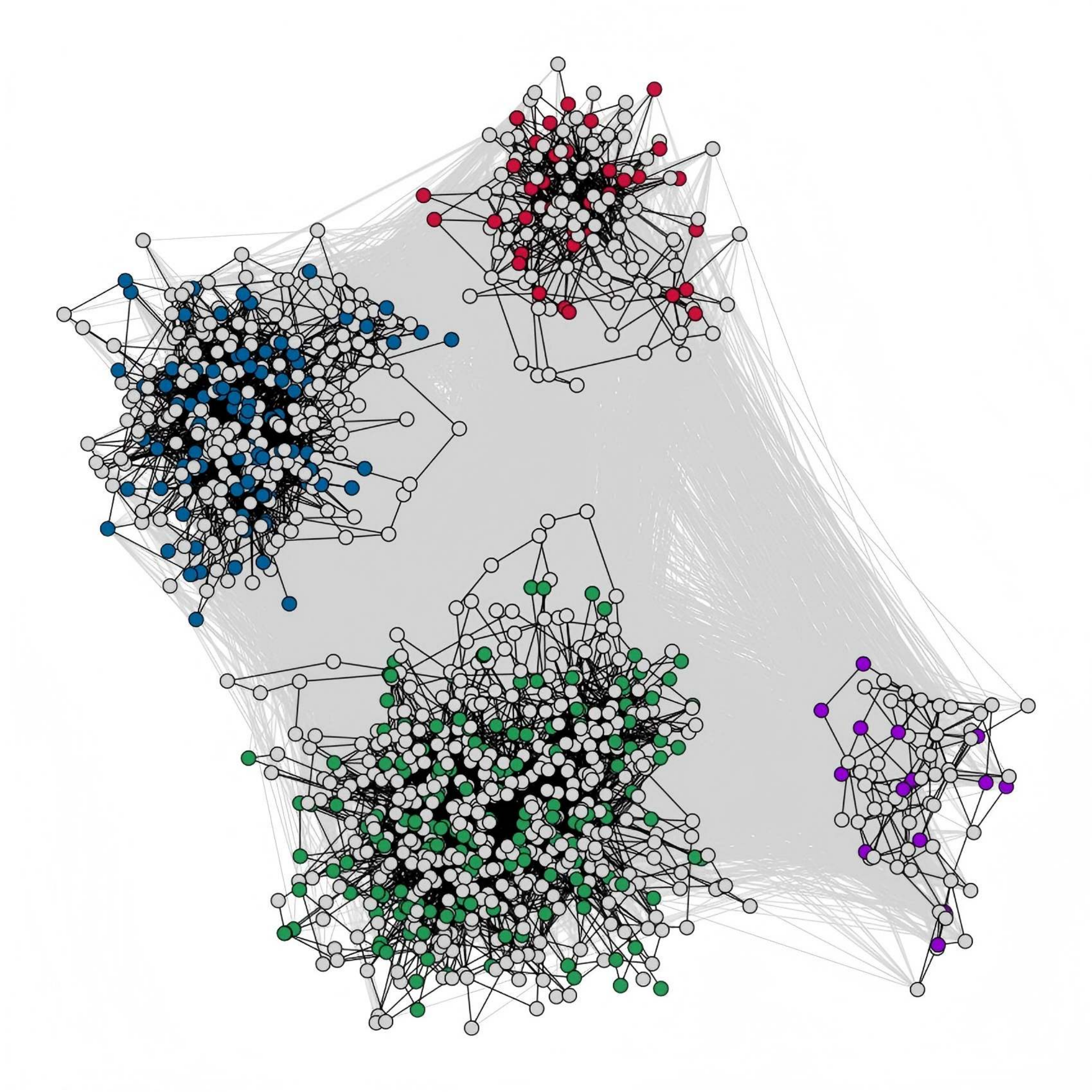}
    }
    \subfigure[GMIM]{
        \includegraphics[width=0.3\linewidth,height=0.3\linewidth]{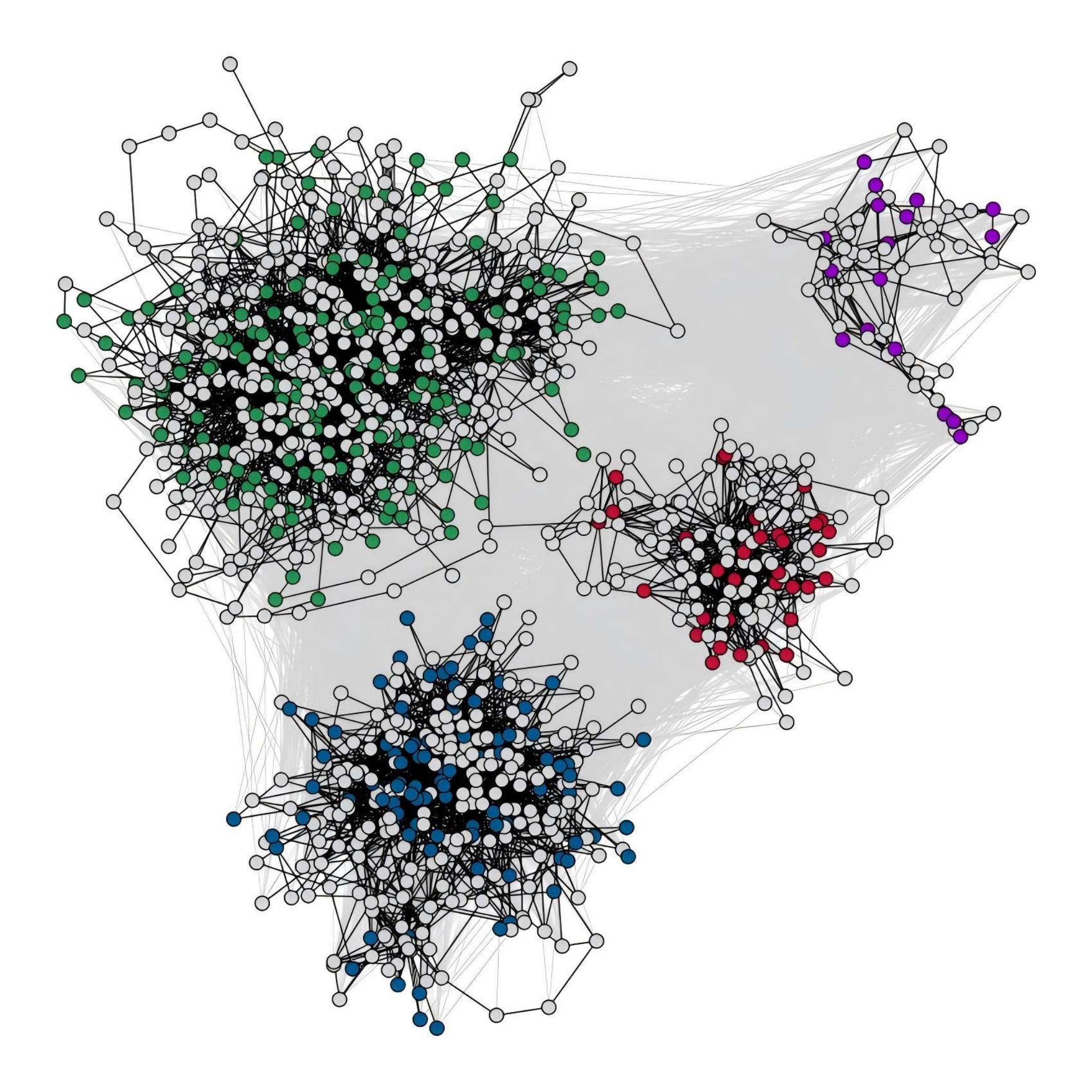}
    }
    \subfigure[DyFSS]{
        \includegraphics[width=0.3\linewidth]{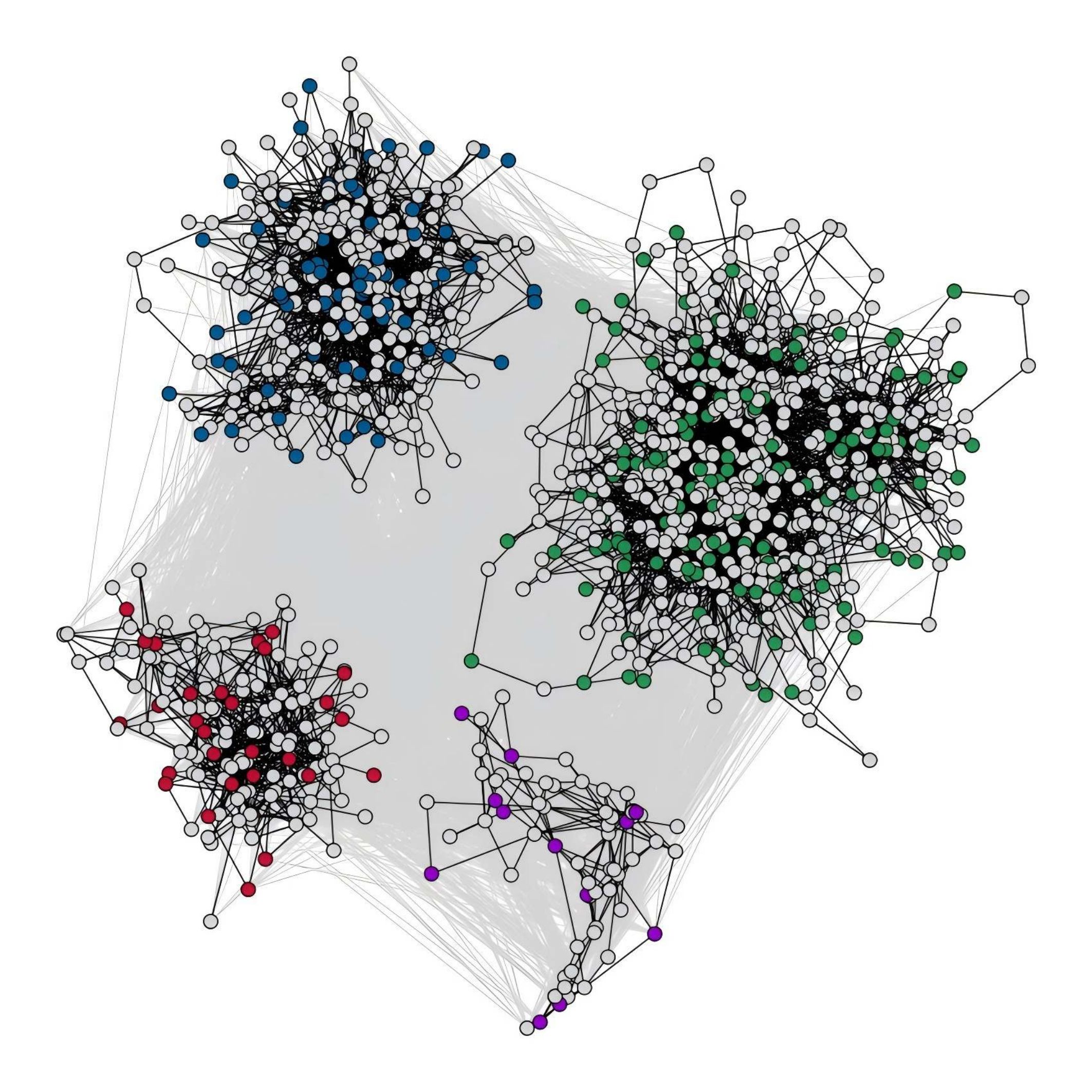}
    }
    \subfigure[CeeGCN]{
        \includegraphics[width=0.3\linewidth,height=0.3\linewidth]{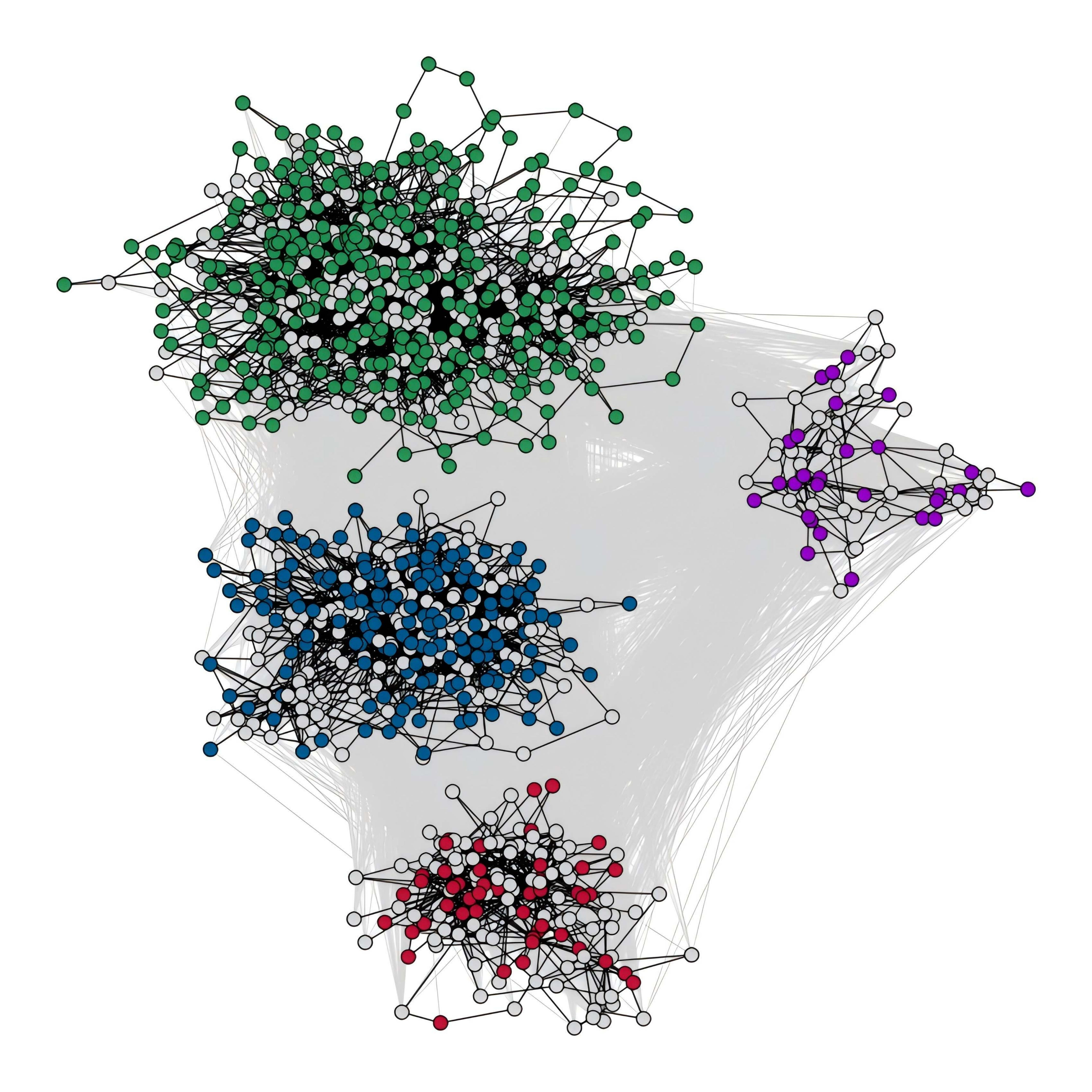}
    }
    \caption{Two-dimensional visualization of clustering results on Vessel01.}
    \label{vis_vessel}
\end{figure}

\subsection{Visualization}
To intuitively compare clustering results across different methods, we visualize the Vessel01 dataset with ground truth labels, where each color represents a distinct cluster. Nodes with inconsistent labels are shown in gray. Thus, a higher proportion of gray nodes in the clustering result indicates poorer performance.

The number of gray nodes in Figure~\ref{vis_vessel} highlights that the clustering result obtained by CeeGCN involve fewer gray nodes, indicating more accurate node assignments to clusters. In contrast, models like GMIM show a higher prevalence of gray nodes. This demonstrates that CeeGCN outperforms these models, providing more accurate clustering results that align closely with ground truth labels.

\end{document}